\documentclass{ieeeaccess}

\usepackage{amsmath,amssymb,amsfonts}

\usepackage{graphicx}
\usepackage{textcomp}
\usepackage{caption}
\usepackage{algorithm}
\usepackage{algpseudocode}
\usepackage{soul}  
\usepackage{multirow}

\def\BibTeX{{\rm B\kern-.05em{\sc i\kern-.025em b}\kern-.08em
    T\kern-.1667em\lower.7ex\hbox{E}\kern-.125emX}}

\begin{document}

\title{When LLM meets Fuzzy-TOPSIS for Personnel Selection through Automated Profile Analysis}
\author{\uppercase{Shahria Hoque}\authorrefmark{1},\uppercase{Ahmed Akib Jawad Karim\authorrefmark{1}},
\uppercase{Md. Golam Rabiul Alam}.\authorrefmark{1}, and Nirjhar Gope\authorrefmark{1},
}
\address[1]{BRAC University Department of Computer Science and Engineering, Dhaka, Dhaka Division, Bangladesh}


\markboth
{Author \headeretal: Preparation of Papers for IEEE TRANSACTIONS and JOURNALS}
{Author \headeretal: Preparation of Papers for IEEE TRANSACTIONS and JOURNALS}

\corresp{Corresponding author: akibjawaad@gmail.com}

\begin{abstract}
In this highly competitive employment environment, the selection of suitable personnel is essential for organizational success. This study presents an automated personnel selection system that utilizes sophisticated natural language processing (NLP) methods to assess and rank software engineering applicants. A distinctive dataset was created by aggregating LinkedIn profiles that include essential features such as education, work experience, abilities, and self-introduction, further enhanced with expert assessments to function as standards. The research combines large language models (LLMs) with multicriteria decision-making (MCDM) theory to develop the LLM-TOPSIS framework. In this context, we utilized the TOPSIS method enhanced by fuzzy logic (Fuzzy TOPSIS) to address the intrinsic ambiguity and subjectivity in human assessments. We utilized triangular fuzzy numbers (TFNs) to describe criteria weights and scores, thereby addressing the ambiguity frequently encountered in candidate evaluations. For candidate ranking, the DistilRoBERTa model was fine-tuned and integrated with the fuzzy TOPSIS method, achieving rankings closely aligned with human expert evaluations and attaining an accuracy of up to 91\% for the \textit{Experience} attribute and the \textit{Overall} attribute. The study underlines the potential of NLP-driven frameworks to improve recruitment procedures by boosting scalability, consistency, and minimizing prejudice. Future endeavors will concentrate on augmenting the dataset, enhancing model interpretability, and verifying the system in actual recruitment scenarios to better evaluate its practical applicability. This research highlights the intriguing potential of merging NLP with fuzzy decision-making methods in personnel selection, enabling scalable and unbiased solutions to recruitment difficulties.
\end{abstract}


\titlepgskip=-15pt
\doi{}

\maketitle

\section{Introduction}
\label{sec:introduction}

\PARstart{R}{ecruiting} capable personnel is a pivotal factor in organizational success, particularly in rapidly evolving technology-driven industries. This study proposes an innovative approach that uses artificial intelligence (AI) and decision-making theories to automate and enhance personnel selection. In today’s dynamic business environment, talent is a key driver of innovation and competitive advantage. The software engineering sector, in particular, demands candidates with specialized skills such as programming expertise, analytical thinking, and domain knowledge. Effective recruitment requires not only attracting suitable candidates but also accurately assessing and ranking them to align with organizational goals. Traditional recruitment methods heavily rely on subjective human judgment, which often leads to inconsistencies, biases, and inefficiencies. This motivates the exploration of automated, data-driven methods to streamline and improve recruitment outcomes. Despite advances in AI-driven recruitment tools, existing solutions typically analyze limited candidate attributes and do not adequately account for the inherent uncertainty and subjectivity in human evaluations. Factors like technical expertise, education, work experience, and self-introduction are complex and multifaceted, making holistic and unbiased candidate appraisal challenging. Consequently, there is a pressing need for an intelligent system that integrates comprehensive candidate analysis with consistent, interpretable, and unbiased decision-making. Current recruitment technologies predominantly employ standalone AI models optimized for automation, often sacrificing interpretability and robustness. Although transformer-based NLP models such as BERT and RoBERTa excel at textual analysis, they are seldom combined with decision-making frameworks. Conversely, classical multicriteria decision-making (MCDM) methods like TOPSIS are effective for ranking but typically struggle with subjective and uncertain inputs. Furthermore, the integration of fuzzy logic, which effectively models vagueness in human judgment, and while in recruitment systems remains limited. These gaps suggest an opportunity to synergize NLP techniques with fuzzy MCDM approaches to build scalable, transparent, and reliable recruitment solutions.

This paper's key contributions include:

\begin{itemize}
    \item Development of a unique dataset comprising LinkedIn profiles augmented with expert evaluations, capturing multiple candidate attributes such as skills, experience, education, and self-introduction.
    
    \item A hybrid framework that integrates LLM-generated, aspect-wise proficiency scores from fine-tuned transformer models (DistilRoBERTa) with fuzzy TOPSIS to explicitly address subjectivity and uncertainty in candidate evaluation and produce reliable rankings.
    
    \item Systematic fine-tuning and empirical validation of DistilRoBERTa as a lightweight, scalable, and reusable backbone model suitable for real-world deployment.
    
    \item Comprehensive empirical validation demonstrating high concordance between model-generated rankings and expert judgments, with DistilRoBERTa achieving up to 91\% accuracy on the Experience and the Overall attribute in classification performance.
    
    \item Establishment of a scalable and interpretable AI-driven recruitment framework that integrates natural language processing with fuzzy decision-making, enabling consistent and human-aligned personnel selection.
\end{itemize}

This study establishes a foundation for future research in scalable, unbiased, and interpretable AI-driven recruitment systems.
The remainder of this paper is organized as follows: Section~\ref{sec:related_work} reviews related work in automated personnel selection, NLP-based candidate assessment, and fuzzy decision-making methods. Section~\ref{sec:methodology} describes the proposed methodology, including dataset preparation, the DistilRoBERTa-based multi-class classification framework, and the fuzzy TOPSIS ranking algorithm. Section~\ref{sec:performance_metrics} presents the experimental setup and evaluation metrics. Section~\ref{sec:results} reports the experimental results and compares model performance with human expert assessments. Section~\ref{sec:discussion} provides a comprehensive discussion of the findings and their practical implications. Section~\ref{sec:limitations_future} discusses the limitations of the study and outlines directions for future work. Finally, Section~\ref{sec:conclusion} concludes the paper and summarizes key contributions.

\section{Related Work}
\label{sec:related_work}

This section critically reviews earlier research on personnel selection systems, a domain where automation remains relatively underexplored. Existing studies predominantly rely on manually collected data from candidates' curricula vitae or resumes, focusing on attributes such as experience, age, and skills \cite{Ayub2009PersonnelSM}. Some works have conducted case studies in contexts like smart villages in Cairo, Egypt, applying structured decision-making approaches to personnel selection~\cite{JeniferB}. For instance, the Technique for Order of Preference by Similarity to Ideal Solution (TOPSIS) combined with the neutrosophic Analytic Hierarchy Process (AHP) has been employed to rank candidates based on multiple criteria, reflecting expert judgments~\cite{Nabeeh}.

Further research has utilized tools such as SADGAGE and methods including ELECTRE III, fuzzy logic modules in MATLAB, and neutrosophic AHP with TOPSIS to address the complexity of human resource selection~\cite{Polychroniou}. Expert input has been sought through interviews with HR managers across various sectors, such as maritime companies in Greece, to enhance decision quality~\cite{koutra2017multicriteria}. The Analytic Network Process (ANP), with the addition of the fuzzy concept, has been proposed to improve reliability by capturing interdependencies among criteria and fostering expert consensus~\cite{Ayub2009PersonnelSM}. Despite these advancements, the majority of these approaches lack full automation and depend heavily on manual assessments.

In the context of automated personnel selection, the foundational work by Karim et al.~\cite{karim2024automated} presents an automated system for evaluating software engineering candidates using large language models (LLMs). This IEEE conference paper introduced the initial LLM-based profile evaluation framework, leveraging transformer models for candidate assessment. The current journal article significantly extends this work by incorporating fuzzy TOPSIS decision-making to better handle subjectivity and uncertainty inherent in human evaluations, thereby enhancing robustness and interpretability.

Our approach develops a fully automated personnel selection system leveraging data collected from LinkedIn profiles, encompassing Skills, Experience, Education, and candidate self-introductions. These data were supplemented with expert evaluations to generate rankings for senior software engineering positions. Central to our approach is the use of sentiment analysis integrated with the RoBERTa model, a transformer-based NLP architecture renowned for its deep self-attention mechanisms and ability to model long-range dependencies in textual data~\cite{liu2019robertarobustlyoptimizedbert}. Additionally, distilled transformer models such as DistilRoBERTa~\cite{sanh2019distilbert} and LastBERT~\cite{karim2025largermodels} have demonstrated the feasibility of creating lightweight yet high-performing models through knowledge distillation, which is particularly beneficial for resource-efficient automated personnel evaluation.

The TOPSIS methodology, widely recognized in decision-making research, structures the evaluation by constructing a decision matrix quantifying how well alternatives satisfy criteria, followed by weighting, normalization, and calculation of distances to ideal best and worst solutions~\cite{Nabeeh}. Our adaptation includes the fuzzy TOPSIS variant, which incorporates triangular fuzzy numbers to model the uncertainty and subjectivity prevalent in human evaluations, thereby enhancing the robustness of candidate rankings~\cite{karsak2002fuzzy}.

Building on the Attraction-Selection-Attrition (ASA) framework, the Person-Environment (P-E) Fit perspective emphasizes multifaceted compatibility between candidates and organizations. In this study, we operationalize this compatibility across individual (Person-Job), team (Person-Group), and organizational (Person-Organization) dimensions, utilizing Large Language Models (LLMs) and Fuzzy TOPSIS to model these complex, non-linear growth dynamics.~\cite{kristof1996person}. These models provide a theoretical basis for understanding personnel selection as a dynamic process influenced by both competencies and cultural fit~\cite{shipp2011coming}. Concurrently, the use of social networking platforms like LinkedIn has gained prominence in recruitment, providing rich candidate data and enabling more efficient selection processes~\cite{melao2020using}.

Although transformer models such as DistilRoBERTa have been successfully applied in specialized NLP tasks like tax law classification and sentiment analysis~\cite{9745941, prytula3fine}, their application in personnel selection remains nascent. While diverse machine learning approaches have been deployed in fields such as sports analytics~\cite{Tulabandhula, Sankaranarayanan, Ganeshapillai} and finance~\cite{yang2020finbert}, these works are largely tangential to recruitment.

Foundational NLP architectures such as BERT and Transformer underpin the current advances in semantic analysis, with variants like Sentence-BERT (SBERT) facilitating efficient transfer learning for tasks including text clustering and classification~\cite{liu2019robertarobustlyoptimizedbert, vaswani2023attentionneed, ORTAKCI2024101730}. Our automated selection framework synthesizes these advances by integrating transformer-derived embeddings with fuzzy decision-making to provide scalable, consistent, and interpretable candidate assessments.

Given the breadth and depth of existing literature, our contribution lies in bridging these domains to construct a robust, automated personnel selection system that leverages state-of-the-art NLP and fuzzy MCDM methods applied to real-world LinkedIn profile data enriched with expert ratings. This system holds promise for applications beyond software engineering recruitment, including defense, healthcare, education, and finance, wherever complex multi-criteria decisions under uncertainty prevail.

\section{METHODOLOGY}
\label{sec:methodology}

This section describes the overall methodology adopted to develop an automated personnel selection framework based on transformer-based language models and fuzzy multi-criteria decision-making. The proposed system consists of three main phases:

\begin{itemize}
    \item \textbf{Dataset Construction and Expert Annotation:}
    A labeled dataset of candidate profiles was constructed from LinkedIn data, covering four key attributes: \textit{Experience}, \textit{Skills}, \textit{Education}, and \textit{About}. Each attribute was evaluated by domain experts using a standardized five-point Likert-scale rubric, which was subsequently mapped to three proficiency classes (\textit{Poor}, \textit{Fair}, \textit{Excellent}). To address data scarcity and class imbalance, the dataset was augmented using paraphrasing, synonym substitution, and contextual expansion, followed by class-wise balancing.

    \item \textbf{Transformer-Based Multi-Class Classification:}
    The DistilRoBERTa model was fine-tuned to perform multi-class classification for each candidate attribute. The model predicts proficiency levels based on textual inputs, producing stable and interpretable attribute-wise scores. Comparative experiments with a larger model (RoBERTa-base) and a smaller model (LastBERT) were also conducted to evaluate the effectiveness of lightweight distilled architectures under limited-data settings.

    \item \textbf{LLM-Generated Proficiency Scores with Fuzzy-TOPSIS-Based Ranking:}

    To aggregate attribute-level predictions into a final candidate ranking, the Fuzzy Technique for Order of Preference by Similarity to Ideal Solution (Fuzzy TOPSIS) was applied. Expert-informed criterion weights were modeled using triangular fuzzy numbers to capture uncertainty and subjectivity. Rankings generated using DistilRoBERTa-derived scores were compared against Human-TOPSIS and direct human expert rankings to assess alignment, robustness, and decision consistency.
\end{itemize}

\subsection{Dataset Construction and Expert Annotation}

A robust dataset forms the backbone of any machine-learning effort, enabling effective training and evaluation of models. In this work, a dataset was meticulously selected and processed to ensure relevance and consistency for automated staff selection in the software engineering area.

The dataset was produced from 100 LinkedIn profiles of software developers. Profiles were selected based on their completeness and relevance to the study, concentrating on the following essential attributes:
Experience: Years worked at numerous companies, demonstrating professional skills.
Education: Details about degrees, certifications, and graduation years.
Skills: A list of programming languages, frameworks, and tools demonstrating technical expertise.
About Section: Self-written descriptions offering insights into personal attributes, professional aspirations, and interpersonal skills.
To offer a benchmark for model training, three senior industry professionals evaluated each profile and provided an Overall Score (1–5) based on the rubric (Appendix A) \label{appendix:rubric}. The dataset was separated into five sub-datasets, each reflecting a unique attribute (e.g., Skills, Experience, Education, About, and Overall) with the score of every candidate.

\subsubsection{Data Description}
The dataset consisted of textual data linked with matching scores, offering a dual framework for evaluation. Here is the main Dataset structure:
After collecting the software engineer’s profile data, we sent that information to a few senior experts. They had evaluated and had put their assessment scores on“Experience”, “Education”, “Skills”, “About”, and “Overall”. Candidate evaluations were performed by experienced domain experts with professional expertise in software engineering recruitment, ensuring reliability and consistency in the scoring process. The Table~\ref{table:back} and Table~\ref{table:evaluation_scores_formatted} represent our data and labeling. 
\begin{table}[h!]
\caption{Professional Background of Candidates}
\centering
\resizebox{0.49\textwidth}{!}{%
\begin{tabular}{|p{1.3cm}|p{1.7cm}|p{1.3cm}|p{1.5cm}|p{1.3cm}|}
\hline
\textbf{Candidates} & \textbf{Current Role} & \textbf{Experience} & \textbf{Education} & \textbf{Skills} \\
\hline
Raju Ahmed
 & Full Stack Eng., Tixio.io & 11y 2m & CUET & Node.js, Laravel, OAuth, React.js, ... \\
\hline
Ibrahim Chowdhury
 & Sr. SW Eng., BJIT & 5y 1m & NSU & Java, Spring Boot, AWS, Git, ... \\
\hline
Anisur Rahman
 & SW Eng., Kinetik & 2y 8m & SUST & Go, AWS, Redis, RabbitMQ, Docker, ... \\
\hline
Md Ferdous 
 & Lead Eng., Banglalink & 8y 5m & DIU & Java, Spring, ML, Angular, Hibernate, ... \\
\hline
Nahid Hasan
 & Assoc. SW Eng., Field Nation & 2y 7m & RU & React, Node.js, K8s, Redis, Next.js, ... \\
\hline
\end{tabular}
}
\label{table:back}
\end{table}

\vspace{1em}

\begin{table}[h!]
\caption{Evaluation Scores of Candidates (1 = Poor, 5 = Outstanding)}
\centering
\resizebox{0.48\textwidth}{!}{%
\begin{tabular}{|p{1.6cm}|p{1.5cm}|p{1.4cm}|p{0.7cm}|p{0.8cm}|p{0.9cm}|}
\hline
\textbf{Candidate} & \textbf{Experience} & \textbf{Education} & \textbf{Skills} & \textbf{About} & \textbf{Overall} \\
\hline
Raju Ahmed & 4 & 3 & 3 & 2 & 3 \\
\hline
Ibrahim Chowdhury & 4 & 3 & 4 & 2 & 3 \\
\hline
Anisur Rahman  & 2 & 4 & 2 & 2 & 2 \\
\hline
Md Ferdous  & 4 & 4 & 3 & 3 & 3 \\
\hline
Nahid Hasan & 3 & 3 & 3 & 2 & 3 \\
\hline
\end{tabular}
}
\label{table:evaluation_scores_formatted}
\end{table}

\begin{figure}[htp]
    \centering
    \includegraphics[width=8cm]{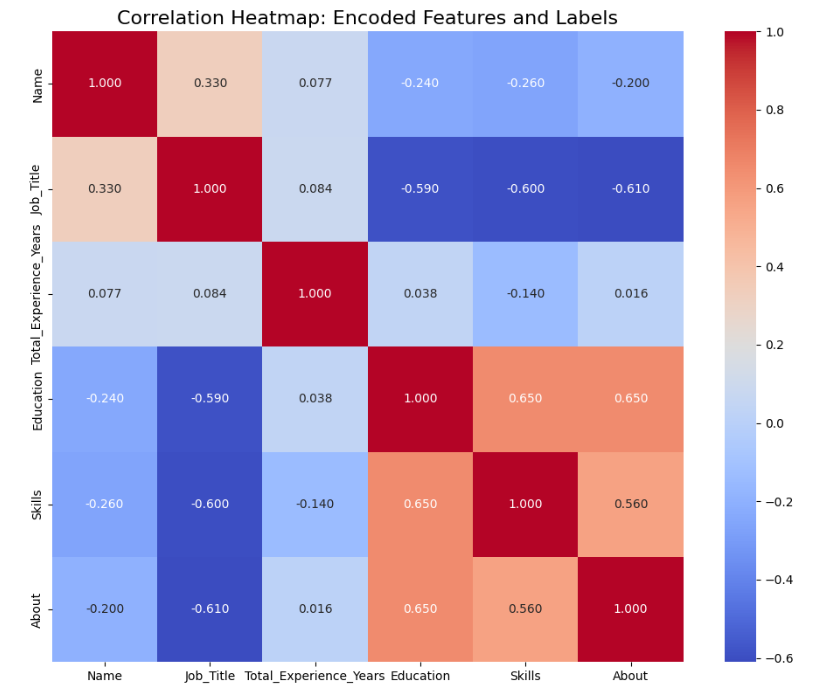}
    \caption{Correlation Heatmap: Encoded Features and Labels}
    \label{fig:Figure1}
\end{figure}

Each sub-dataset had: A textual input column, including processed text for each attribute. A numerical score column, representing expert evaluations.
A total of 10,000 samples were created for each sub-dataset to provide appropriate data diversity by using augmentation techniques via synonym substitution, paraphrasing, and contextual expansion. The correlation analysis indicates important correlations between the encoded properties, as seen in the heatmap (Figure~\ref{fig:Figure1}). A strong positive correlation (0.65) exists between ‘Education’ and ‘Skills’, showing that individuals with greater educational levels frequently possess more extensive skill sets. Similarly, the ‘About’ feature shows a positive connection with both ‘Education’ (0.65) and ‘Skills’ (0.56), demonstrating that well-crafted biographical descriptions are connected with greater qualifications and broader skill sets. 

Interestingly, there are traits with neutral or weak associations. For instance, ‘Total Experience Years’ shows a negligible association with ‘Education’ (0.038) and a weak negative correlation with ‘Skills’ (-0.14). This pattern implies that experience alone may not necessarily match closely with higher education or specific talents.

On the other hand, ‘Job Title’ demonstrates a negative connection with both ‘Education’ (-0.59) and ‘talents’ (-0.60), which would signal that persons in senior posts may prefer experience over formal qualifications or diverse talents.

These observations, covering strong, neutral, and negative associations, led to the prioritizing of characteristics for the future modeling phase. Specifically, the weak correlations between experience and other attributes support treating Experience, Skills, Education, and About as independent criteria within the proposed multi-criteria decision-making framework.

Moreover, the original dataset exhibited a slight imbalance across the three classes (Poor, Fair, Excellent), with the Fair category being less represented. Random oversampling was applied within each sub-dataset to achieve balanced proportions across all labels. Table 3 \label{tab:class_distribution} summarizes the class distributions before and after balancing. This ensured that no single class dominated the training process, contributing to the improved recall of the Fair category observed in later evaluations.

\begin{table}[h!]
\centering
\caption{Class Distribution Before and After Balancing Across Sub-Datasets}
\begin{tabular}{|l|c|c|c|c|c|c|}
\hline
\multirow{2}{*}{\textbf{Sub-Dataset}} & \multicolumn{3}{c|}{\textbf{Before Balancing (\%)}} & \multicolumn{3}{c|}{\textbf{After Balancing (\%)}} \\ \cline{2-7} 
 & Poor & Fair & Excellent & Poor & Fair & Excellent \\ \hline
Experience & 39 & 18 & 43 & 33.3 & 33.3 & 33.3 \\ \hline
Skills & 37 & 22 & 41 & 33.3 & 33.3 & 33.3 \\ \hline
Education & 40 & 19 & 41 & 33.3 & 33.3 & 33.3 \\ \hline
About & 38 & 20 & 42 & 33.3 & 33.3 & 33.3 \\ \hline
Overall & 36 & 21 & 43 & 33.3 & 33.3 & 33.3 \\ \hline
\end{tabular}
\label{tab:class_distribution}
\end{table}

To prepare the dataset for machine learning, the following preprocessing processes were performed:
Text Cleaning and Normalization: Text was converted to lowercase, punctuation and special symbols were deleted, and multi-space characters were replaced with a single space. Synonyms were substituted using WordNet to promote textual diversity.
Proficiency score Label Mapping: Scores were divided into labels—Poor (1-2), Fair (3), and Excellent(4-5) for supervised training.
Data Augmentation: Synonym-enhanced text samples were added to the original data, thereby doubling the dataset size.
Class Balancing: Resampling approaches provided equal representation of all feeling categories.
Dataset Splitting: The dataset was split into training, validation, and test sets (80-10-10) to ensure robust evaluation.
By utilizing these methodologies, the dataset was modified to address difficulties such as data imbalance and limited diversity, giving a solid platform for model development and validation.


\subsection{DistilRoBERTa-Based Scoring with Fuzzy TOPSIS Aggregation for Multi-Criteria Candidate Evaluation} 

\begin{figure}[htbp]
    \centering
    \includegraphics[width=0.5\textwidth]{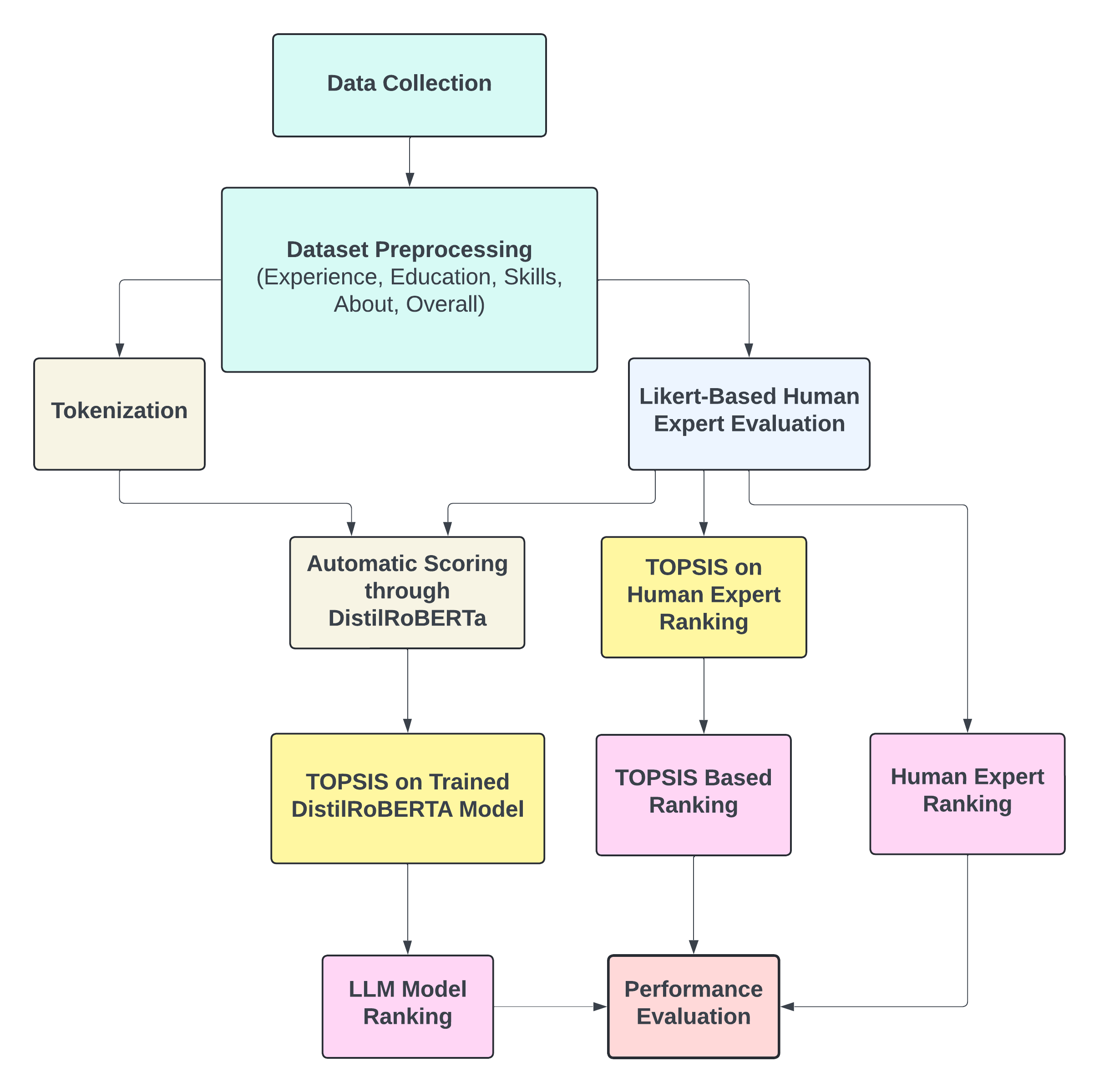}
    \caption{Top-level overview of the LLM-Fuzzy TOPSIS Integration for personal selection}
    \label{fig:workflow}
\end{figure}


The Figure \ref{fig:workflow} and the Algorithm \ref{algorithm:1} showcase the top-level overview of this proposed method, where DistilRoBERTa is a compact variant of the RoBERTa (Robustly Optimized BERT Pretraining Approach) language model, developed for efficiency while maintaining commendable performance. These large language models (LLMs) were chosen for their robust contextual comprehension and capacity to manage complicated textual material while ensuring computational efficiency. It is instructed by distillation of data, achieving around 97\% of the complete RoBERTa model's performance while being 40\% smaller and 60\% faster. DistilRoBERTa integrates transformer architecture with attention mechanisms to analyze text, rendering it very effective for natural language comprehension tasks, including classification, ranking, and sentiment analysis. Its shortened size and computational demands render it suitable for resource-limited applications while maintaining accuracy substantially.

\begin{algorithm}
\scriptsize
\caption{DistilRoBERTa + Fuzzy TOPSIS for Candidate Ranking}
\label{algorithm:1}
\begin{algorithmic}[1] 
\State \textbf{Input:} Candidate dataset $D = \{x_i\}_{i=1}^N$ with text features (Skills, Experience, Education, About)
\State \textbf{Input:} Label categories $\mathcal{Y} = \{\text{Poor}, \text{Fair}, \text{Excellent}\}$
\State \textbf{Input:} Trained DistilRoBERTa models $\mathcal{M}_{skill}, \mathcal{M}_{exp}, \mathcal{M}_{edu}, \mathcal{M}_{about}$
\State \textbf{Output:} Ranked list of candidates

\Procedure{RankCandidates}{$D, \mathcal{M}_{skill}, \mathcal{M}_{exp}, \mathcal{M}_{edu}, \mathcal{M}_{about}$}
    \For{each candidate $x_i \in D$}
        \State Tokenize all four text fields with RoBERTa tokenizer (max length = 256)
        \State Predict labels $y_i^{(skill)} = \mathcal{M}_{skill}(x_i^{(skill)})$
        \State Predict labels $y_i^{(exp)} = \mathcal{M}_{exp}(x_i^{(exp)})$
        \State Predict labels $y_i^{(edu)} = \mathcal{M}_{edu}(x_i^{(edu)})$
        \State Predict labels $y_i^{(about)} = \mathcal{M}_{about}(x_i^{(about)})$
        \State Map predicted labels to scores: \textit{Poor} $\rightarrow$ 1--2, \textit{Fair} $\rightarrow$ 3, \textit{Excellent} $\rightarrow$ 4--5
        \State Form decision vector $v_i = \left[s_i^{\text{(skill)}},\ s_i^{\text{(exp)}},\ s_i^{\text{(edu)}},\ s_i^{\text{(about)}}\right]$

    \EndFor

    \State Form decision matrix $V = [v_1; v_2; \dots; v_N]$
    \State \textbf{// Fuzzy TOPSIS Steps}
    \State Normalize decision matrix $V$
    \State Apply fuzzy weights using Triangular Fuzzy Numbers (TFNs)
    \State Identify FPIS and FNIS
    \For{each candidate $v_i$}
        \State Compute distance to FPIS: $D_i^+$
        \State Compute distance to FNIS: $D_i^-$
        \State Compute closeness coefficient: $CC_i = \frac{D_i^-}{D_i^+ + D_i^-}$
    \EndFor

    \State Rank candidates by descending $CC_i$
    \State \Return ranked list
\EndProcedure
\end{algorithmic}
\end{algorithm}

\begin{figure}[htbp]
    \centering
    \includegraphics[width=0.5\textwidth]{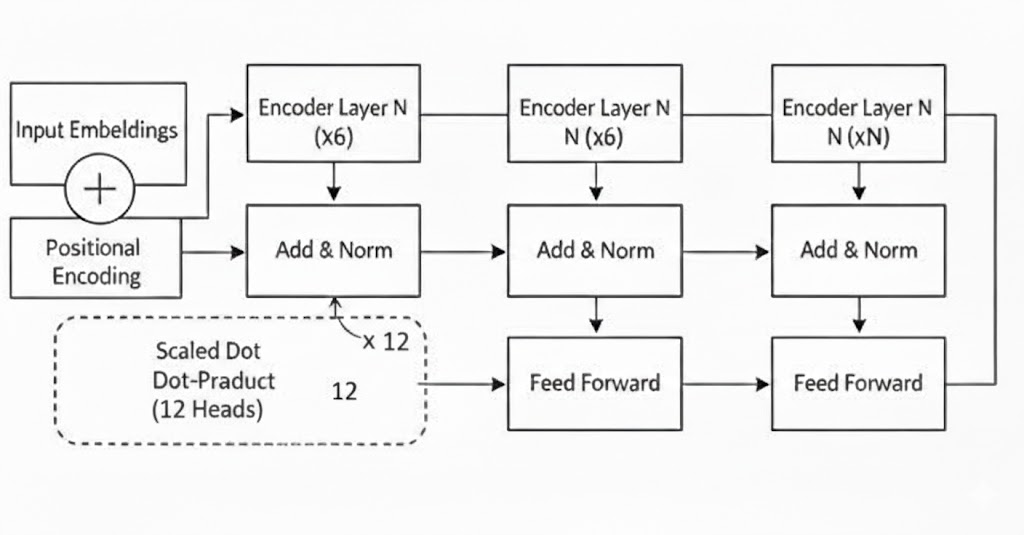}
    \caption{Top-level overview of the DistilRoBERTa's internals}
    \label{fig:top-level-1}
\end{figure}

\begin{figure}[htbp]
    \centering
    \includegraphics[width=0.5\textwidth]{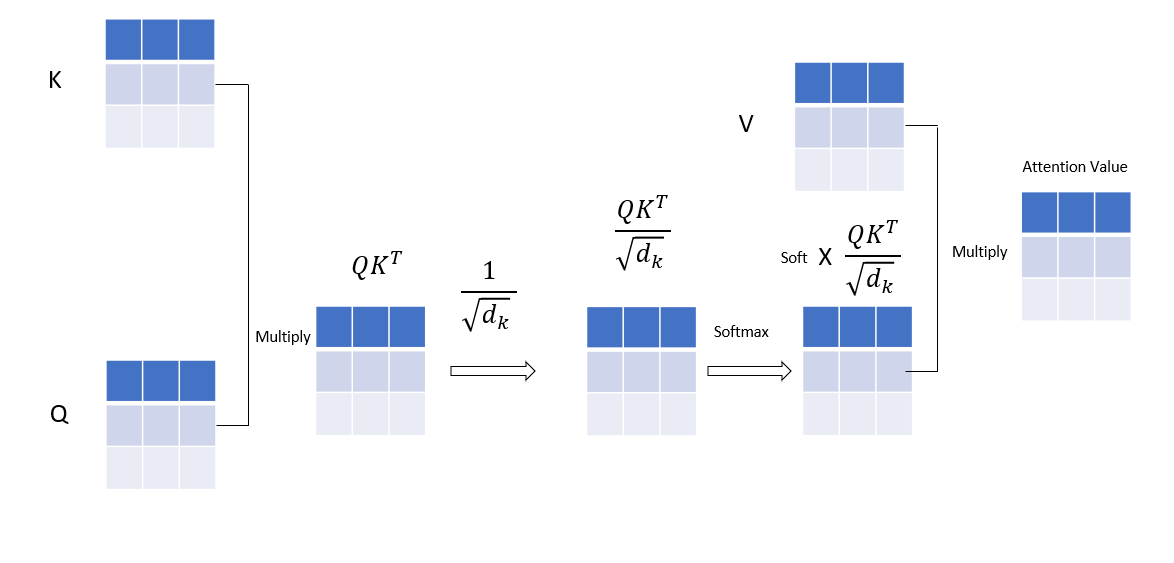}
    \caption{Top-level overview of the Attention head}
    \label{fig:top-level-2}
\end{figure}

Figures \ref{fig:top-level-1}, \ref{fig:top-level-2} explains the internal aspects step by step in detail. DistilRoBERTa is based on the transformer architecture and achieves comparable performance to RoBERTa with fewer parameters and faster computation. The DistilRoberta Base model has 6 layers, 768 dimensions and 12 heads, totaling 82M parameters (compared to 125M parameters for RoBERTa-base). On average, DistilRoBERTa is twice as fast as Roberta-base.

\subsubsection{Model Training and Hyperparameter Settings}

Table~\ref{tab:hyperparams} summarizes the architecture and training hyperparameters used for all evaluated transformer models. To ensure a fair comparison, each model was trained using task-appropriate configurations, with learning rates and batch sizes selected according to model capacity and stability requirements. A linear learning-rate warmup strategy was applied, with a warmup ratio of 6\% for LastBERT and 10\% for both RoBERTa-base and DistilRoBERTa-base, followed by cosine decay. All models were fine-tuned on the same augmented dataset and selected based on the lowest validation loss to prevent overfitting.

\begin{table}[htbp]
\centering
\caption{Model Architectures and Training Hyperparameters}
\renewcommand{\arraystretch}{1.15}
\setlength{\tabcolsep}{4pt}
\begin{tabular}{|l|c|c|c|c|c|}
\hline
\textbf{Model} &
\textbf{Epochs} &
\textbf{LR} &
\textbf{Batch} &
\textbf{Max Seq.} \\ \hline

LastBERT
& 20 & $1\times10^{-7}$ & 8 & 256 \\ \hline

RoBERTa-base
& 18 & $2\times10^{-5}$ & 32 & 512 \\ \hline

DistilRoBERTa-base
 & 18 & $1\times10^{-5}$ & 16 & 256 \\ \hline

\end{tabular}
\label{tab:hyperparams}
\end{table}

\subsubsection{Architecture of DistilRoBERTa and Scaled Dot-Product Attention Mechanism}

The input sequence, typically consisting of text tokens, is first converted into embeddings. Each token \( x_i \) is mapped to a dense vector representation using a learned embedding layer. Since transformer-based models do not inherently encode the positional order of tokens, positional encodings are added to these embeddings to provide sequential context.

The core component of DistilRoBERTa is its encoder block, which is repeated six times, unlike the original RoBERTa model, which uses twelve layers. Each encoder layer comprises a multi-head self-attention mechanism followed by a position-wise feed-forward network (FFN). Residual connections and layer normalization are employed after both the attention and FFN sublayers to enhance training stability and convergence.

The self-attention mechanism enables the model to attend to different parts of the sequence simultaneously, learning contextual relationships across tokens. The attention output is passed through the FFN, allowing the model to capture non-linear transformations and higher-level semantic representations. After the final encoder layer, the hidden representation is projected via a linear transformation into logits, which are then passed through a softmax layer to produce probability distributions over target classes. This setup is particularly effective for classification tasks such as next-word prediction or candidate proficiency scoring.

\subsubsection{Mathematical Representation of the Model Components}

\paragraph*{Input Representation:}
Given an input token sequence \( X = [x_1, x_2, \dots, x_n] \), each token \( x_i \) is transformed into a vector embedding:
\begin{equation}
    e(x_i) = \text{Embedding}(x_i)
\end{equation}
These embeddings are combined with positional encodings to account for token order, forming the input to the encoder.

\paragraph*{Scaled Dot-Product Attention:}
The self-attention mechanism calculates attention weights using the scaled dot-product formula:
\begin{equation}
\text{Attention}(Q, K, V) = \text{softmax}\left(\frac{QK^\top}{\sqrt{d_k}}\right)V
\end{equation}
Here, \( Q \), \( K \), and \( V \) denote the query, key, and value matrices, respectively, and \( d_k \) is the dimensionality of the key vectors.

\paragraph*{Multi-Head Attention:}
To allow the model to jointly attend to information from different representation subspaces, multiple attention heads are employed:
\begin{equation}
\text{MultiHead}(Q, K, V) = \text{Concat}(head_1, \dots, head_h) W^O
\end{equation}
where each head independently computes attention, and the outputs are concatenated and linearly transformed via \( W^O \).

\paragraph*{Feed-Forward Network (FFN):}
Each encoder block includes a position-wise feed-forward neural network defined as:
\begin{equation}
\text{FFN}(x) = \text{ReLU}(xW_1 + b_1)W_2 + b_2
\end{equation}
where \( W_1 \), \( W_2 \) are learnable weight matrices and \( b_1 \), \( b_2 \) are bias vectors. The FFN applies a non-linear transformation to each position independently.

\subsubsection{Training Objective}

DistilRoBERTa is fine-tuned for the task of applicant profiling, where the goal is to classify and rank candidates based on their proficiency scores. The overall training loss combines standard cross-entropy loss with a knowledge distillation component to ensure that the student model mimics the behavior of a larger, pre-trained teacher model. The loss function is expressed as:
\begin{equation}
\mathcal{L}_{\text{distill}} = \lambda \cdot \mathcal{L}_{\text{CE}} + (1 - \lambda) \cdot \mathcal{L}_{\text{KD}}
\end{equation}
where:
\begin{itemize}
    \item \( \mathcal{L}_{\text{CE}} \) is the cross-entropy loss between predicted and true class labels.
    \item \( \mathcal{L}_{\text{KD}} \) is the Kullback–Leibler divergence between the teacher's and student's predicted distributions:
\end{itemize}
\begin{equation}
\mathcal{L}_{\text{KD}} = \text{KL}\left(p_{\text{teacher}}(z) \parallel p_{\text{student}}(z)\right)
\end{equation}
The hyperparameter \( \lambda \in [0,1] \) governs the trade-off between direct supervision and knowledge transfer.

\subsubsection{Final Output and Prediction}

The final hidden representation produced by the last encoder layer is passed through a linear projection layer, followed by a softmax activation to generate class probabilities:
\begin{equation}
\mathbf{y} = \text{Softmax}(W_{\text{out}} \cdot \text{Output})
\end{equation}
where \( W_{\text{out}} \) is the output weight matrix. The softmax function normalizes the logits to produce a probability distribution over the candidate proficiency categories. These probabilities serve as the basis for downstream decision-making, including ranking or classification of applicants.

The multi-head attention technique in Figure \ref{fig:top-level-1} is achieved using the scaled dot-product attention described in Figure \ref{fig:top-level-2}. It allows the model to interpret information from numerous viewpoints by employing multiple attention heads. The DistilRoBERTa model has 6 encoder layers, compared to 12 in the original RoBERTa, making it more efficient while preserving most of the performance. This design enables DistilRoBERTa, a strong yet lightweight model for natural language processing applications.

To develop the DistilRoBERTa-based sequence categorization, numerous libraries were leveraged to provide an effective workflow. The major framework was Hugging Face's Transformers library (version 3.0.2), which contains pre-trained models such as DistilRoBERTa. Additional libraries included pandas and numpy for data management, scikit-learn for dividing datasets, and torch for deep learning operations.
The dataset was loaded from CSV files into pandas DataFrames. A mapping function translated numerical scores into categorical labels—Poor, Fair, and Excellent. This mapping was systematically applied to all columns, such as Skills, Experience, Education, and About, to preserve uniformity in label representation. The modified data was evaluated to validate class balance across the categories.
For text preprocessing, the RobertaTokenizer from Hugging Face was customized with padding, truncation, and a maximum token length of 256. Tokenized outputs were turned into PyTorch tensors, incorporating input IDs and attention masks. The dataset was separated into training (80\%), validation (10\%), and test (10\%) sets using a fixed random seed for reproducibility. The splits were saved as separate CSV files for simple incorporation into the model training workflow.
A new PyTorch Dataset class was designed to manage tokenized data and accompanying labels. It leveraged the RoBERTa tokenizer to transform text inputs into acceptable features for model training. DataLoader objects were developed for batching the data during training and validation, ensuring efficient data processing. Validation of the tokenization procedure was accomplished by inspecting sample batches.
The DistilRobertaClass model, based on DistilBERTForSequenceClassification, was initialized using the distilroberta-base checkpoint. The model architecture was configured for three output classes: Poor, Fair, and Excellent. Input IDs and attention masks drove the model through its embedding layers and transformer blocks to produce logits for classification. Cross-Entropy Loss was applied as the training criterion, and the AdamW optimizer was used with a learning rate of 2e-5. Training was undertaken on a GPU (Tesla T4) to leverage computational performance.
The model was fine-tuned on multiple datasets to predict performance labels based on their individual attributes. Training spanned 18 epochs for the Skills and Experience datasets, 19 epochs for the About dataset, and 15 epochs for the Education dataset. Early stopping was utilized to prevent overfitting, interrupting training if validation loss did not improve for two consecutive epochs.
Each training epoch includes a forward pass through the model to compute loss, followed by backpropagation for weight changes. Validation was undertaken after each epoch to evaluate the model's performance on unseen data. Metrics such as training loss, validation loss, and accuracy were tracked during the procedure. Early stopping ensured efficient training by quitting when performance plateaued.
The completed model was saved, and predictions were created for each dataset column. Scores for individual candidates were calculated and retained for evaluation. This fine-tuning technique, employing DistilRoBERTa, provides a successful and resource-efficient solution for multi-class classification problems, enabling accurate prediction of performance labels across varied features.

\subsubsection{Fuzzy TOPSIS Ranking Algorithm}

Fuzzy TOPSIS (Technique for Order Preference by Similarity to Ideal Solution) is an advanced decision-making system that uses fuzzy logic to manage ambiguity and vagueness in data. It uses triangular fuzzy numbers (TFNs) to describe criteria and weights, allowing for more realistic judgments in subjective circumstances. The algorithm creates a normalized choice matrix, derives the ideal best and worst values, and computes the distance of each alternative from these ideals using fuzzy Euclidean distance. The alternatives are rated based on their relative closeness to the ideal answer, offering a strong and interpretable ranking framework for complex decision scenarios.

The combination of Large Language Models (LLMs) with Fuzzy-TOPSIS is motivated by their complementary roles in automated personnel selection. While LLMs are effective at extracting rich, contextual information from unstructured profile data, their outputs are inherently qualitative and uncertain. Fuzzy-TOPSIS provides a principled mechanism to model this uncertainty through triangular fuzzy numbers and to aggregate multiple evaluation criteria into a single, interpretable ranking. Compared to alternative MCDM techniques, Fuzzy-TOPSIS offers computational simplicity, transparency, and a clear closeness coefficient, making it well-suited for transforming LLM-derived assessments into robust and explainable candidate rankings.

This method incorporated various decision-making criteria into a scalable and transparent framework, enabling effective ranking and selection of candidates for software engineering vacancies.

In this study, criteria weights and scores are represented using triangular fuzzy numbers (TFNs), defined as $\tilde{a} = (l, m, u)$, where $l$, $m$, and $u$ denote the lower, modal, and upper bounds, respectively. TFNs are used to capture the uncertainty and subjectivity inherent in expert judgments during candidate evaluation.
\begin{table}[htbp]
\centering
\caption{Linguistic Terms and Corresponding Triangular Fuzzy Numbers}
\begin{tabular}{|c|c|}
\hline
\textbf{Linguistic Term} & \textbf{TFN $(l,m,u)$} \\ \hline
Very Low & (0.0, 0.1, 0.3) \\ \hline
Low & (0.1, 0.3, 0.5) \\ \hline
Medium & (0.3, 0.5, 0.7) \\ \hline
High & (0.5, 0.7, 0.9) \\ \hline
Very High & (0.7, 0.9, 1.0) \\ \hline
\end{tabular}
\label{tab:tfn_mapping}
\end{table}
Expert evaluations were initially provided using linguistic terms reflecting perceived importance and proficiency levels. These linguistic assessments were subsequently converted into TFNs using the mapping shown in Table~\ref{tab:tfn_mapping}. When multiple expert opinions were available, fuzzy aggregation was performed by averaging the corresponding lower, modal, and upper values.

Although Eqs.~(9–12) are presented in crisp form for readability, all computations are performed on triangular fuzzy numbers. Final candidate scores are obtained via defuzzification using the centroid method.

Defuzzification is performed as $C(\tilde{a}) = (l + m + u)/3$.

To rank and evaluate software engineering candidates fairly, the Fuzzy TOPSIS methodology was applied to a dataset including four essential criteria: Experience, Skills, Education, and About, with weights allocated as Experience (0.2), Skills (0.6), Education (0.15), and About (0.05). This systematic approach ensured fair and comprehensive decision-making. A structured MCDM weighting procedure, inspired by the Analytic Hierarchy Process (AHP), was used to determine their relative importance. Three senior software engineering experts provided pairwise judgments on the importance of the four criteria. The aggregated comparisons were normalized using an AHP-style eigenvector method, resulting in the final weights: Skills = 0.60, Experience = 0.20, Education = 0.15, and About = 0.05. To incorporate uncertainty in expert judgments, the weights were represented using triangular fuzzy numbers (TFNs), consistent with standard fuzzy MCDM practice. This ensures the weighting scheme is methodologically grounded, reproducible, and aligned with established decision-making frameworks, rather than subjective or arbitrary.

The decision matrix (\ref{eq:decision-matrix}) showcases the candidate scores across different criteria.

$$Step\ 1: Decision\ Matrix$$

\begin{equation}
\label{eq:decision-matrix}
D = 
\begin{pmatrix}
4.5 & 3.8 & 4.2 & 5.0 \\
4.0 & 4.5 & 4.8 & 4.6 \\
3.7 & 4.2 & 4.5 & 4.9
\end{pmatrix}
\end{equation}

By scaling scores into a similar range, normalization assures that scores from various criteria (such as skills and experience) are comparable.
Xij: Combined Score for candidate i and criterion j. According to the decision matrix, if the scores for ”Skills” (j) across three candidates are: [4.5, 4.0, 3.7]

$$Step\ 2: Calculate\ Normalised\ Matrix$$

\begin{equation}
   \bar{X}_{i_j} = \frac{X_{i_j}}{\sqrt{\sum_{i=1}^n X_{i_j}^2}}
\end{equation}

$$Step\ 3: Calculate\ Weighted\ Normalised\ Matrix$$
After normalization, the results are weighted according to their given value (the weight for each category). The weight is multiplied by each corresponding normalized score to obtain the weighted normalized matrix (\ref{eq:pythagorean}).

\begin{equation}
   \label{eq:pythagorean}
   V_{i,j} = \bar{X}_{i,j} \times W_j
\end{equation}

$$Step\ 4: Calculate\ The\ Ideal\ Best\ And\ Ideal\ Worst\ Values$$

For benefit criteria, the fuzzy ideal best and ideal worst solutions are defined as:

\begin{equation}
V_j^+ = \max_i \{ V_{i,j} \}, \quad
V_j^- = \min_i \{ V_{i,j} \}
\end{equation}

where $V_{i,j}$ denotes the weighted normalized value of candidate $i$ under criterion $j$.

$$Step\ 5: Calculate\ Euclidean\ Distance\ From\ The\ Ideal\ Best$$
\begin{equation}
S_i^+ = \left[{\sum_{j=1}^m (V_{i_j} - V_j^+)^2}\right]^{0.5} \label{eq:your_label}
\end{equation}

$$Step\ 6: Calculate\ Euclidean\ Distance\ From\ The\ Ideal\ Worst$$
\begin{equation}
S_i^- = \left[{\sum_{j=1}^m (V_{i_j} - V_j^-)^2}\right]^{0.5} \label{eq:your_label}
\end{equation}

$$Step\ 7: Calculate\ Performance\ Score$$

\begin{equation} \label{eq:Pi}
P_i = \frac{S_i^-}{{S_i^+} + {S_i^-}}
\end{equation}

\section{Performance Metrics}
\label{sec:performance_metrics}

After saving my train model, Every Dataset was assessed on the test dataset using measures including accuracy, F1-score, precision, Hamming loss, and recall. A confusion matrix was built to visually demonstrate how efficiently the model categorized Candidate performances. We examine the DistilBERT, RoBERTa, and LastBERT models based on measures including accuracy, F1 score, precision, recall, and optimal thresholds. Performance in sentiment identification is measured by training curves and confusion matrices, presenting a comparative assessment of each model’s aptitude for this job while balancing sensitivity and specificity. In addition, we employ MAP, NDCG, MRR, MAE, and Cosine Similarity to evaluate the model-generated and human-generated Proficiency scores.

A model, system, or process's overall performance, efficacy, and efficiency are assessed using performance metrics, which are quantitative measurements. These metrics are crucial in determining how well a system or model is performing with respect to a given task. In machine learning, performance metrics are used to assess the success of a model in solving a particular problem, whether it is a classification, regression, or clustering task.

Classification metrics are used when the task involves predicting discrete labels or categories, such as binary classification or multi-class classification.

\begin{itemize}
    \item \textbf{Accuracy}: The proportion of accurate forecasts to all forecasts made. 
    \begin{equation}
    \text{Accuracy} = \frac{\text{True Positives} + \text{True Negatives}}{\text{Total Predictions}}
    \end{equation}
    \item \textbf{Precision}: The proportion of accurately forecasted positive observations to all of the predictions. 
    \begin{equation}
    \text{Precision} = \frac{\text{True Positives}}{\text{True Positives} + \text{False Positives}}
    \end{equation}
    \item \textbf{Recall (Sensitivity)}: The proportion of all actual positive observations to accurately projected as positive observations. 
    \begin{equation}
    \text{Recall} = \frac{\text{True Positives}}{\text{True Positives} + \text{False Negatives}}
    \end{equation}
    \item \textbf{F1-Score}: The precision and recall harmonic mean. When there is an imbalance in the distribution of classes, it is helpful.
    \begin{equation}
    \text{F1-Score} = 2 \cdot \frac{\text{Precision} \cdot \text{Recall}}{\text{Precision} + \text{Recall}}
    \end{equation}

\end{itemize}

\begin{itemize}

    \item \textbf{Mean Absolute Error (MAE)}: The mean of the absolute differences between the actual and predicted values.
    \begin{equation}
        \text{MAE} = \frac{\sum_{i=1}^{n} \left| y_i - \hat{y}_i \right|}{n}
    \end{equation}
    
    \item \textbf{Cosine Similarity}: Measures the cosine of the angle between two non-zero vectors in an inner product space, indicating their directional similarity.
    \begin{equation}
        \text{Cosine Similarity} = \frac{\sum_{i=1}^{n} x_i y_i}{\sqrt{\sum_{i=1}^{n} x_i^2} \sqrt{\sum_{i=1}^{n} y_i^2}}
    \end{equation}
    
    \item \textbf{Root Mean Squared Error (RMSE)}: The square root of the average of squared differences between the actual and predicted values.
    \begin{equation}
        \text{RMSE} = \sqrt{\frac{\sum_{i=1}^{n} (y_i - \hat{y}_i)^2}{n}}
    \end{equation}
    
    \item \textbf{Mean Average Precision (MAP)}: The mean of the average precision scores across all queries.
    \begin{equation}
        \text{MAP} = \frac{1}{Q} \sum_{q=1}^{Q} \text{Average Precision}(q)
    \end{equation}
    
    \item \textbf{Mean Reciprocal Rank (MRR)}: The average of the reciprocal ranks of the first relevant result for a set of queries.
    \begin{equation}
        \text{MRR} = \frac{1}{|Q|} \sum_{i=1}^{|Q|} \frac{1}{\text{rank}_i}
    \end{equation}
    
    \item \textbf{Normalized Discounted Cumulative Gain (NDCG)}: A ranking quality metric that considers the position of relevant results in the ranking.
    \begin{equation}
        \text{NDCG} = \frac{DCG}{IDCG}
    \end{equation}
    
    \item \textbf{Hamming Loss}: The total number of incorrectly predicted labels (either missed positives or falsely added ones) across all candidates.
    
\end{itemize}

Total Labels: The total number of labels across all candidates in the dataset. 
    \begin{equation}
    \text{Hamming Loss} = \frac{\text{Number of Incorrect Labels} }{\text{Total labels}}
    \end{equation}

Number of Incorrect Labels: The total number of labels (for all candidates) that were incorrectly predicted (either missed positive labels or added incorrect positive labels).
Total Labels: The total number of labels across all candidates in the dataset.

\section{Result Analysis}
\label{sec:results}

After training, the saved model was evaluated on the test dataset using standard performance metrics, including accuracy, precision, recall, F1-score, and Hamming loss. A confusion matrix was also generated to visualize the classification performance in detail.

The large language model's effectiveness was assessed through these metrics, along with an analysis of optimal decision thresholds. Training and validation curves were reviewed to evaluate learning behavior and identify signs of overfitting or underfitting. The confusion matrix provided additional insight into the model’s ability to accurately classify candidate performance levels, highlighting the balance achieved between sensitivity and specificity during sentiment classification.

\subsection{Performance of DistilRoBERTa on Experience Attribute}

The DistilRoBERTa model displays high capabilities in sentiment categorization, shown in Table~\ref{tab:experience_performance}. It performs exceptionally in the "Poor" category, with a precision of 0.95, a recall of 1.0, and an F1-score of 0.97. For "Fair," although the precision is excellent (1.0), recall reduces to 0.36, leading to a lower F1-score of 0.53. The "Excellent" category similarly displays strong results, with a precision of 0.89, a recall of 0.99, and an F1-score of 0.94. The overall accuracy stands around 91\%, indicating its reliable performance across numerous emotion categories.

\begin{table}[h]
\centering
\caption{PERFORMANCE METRICS FOR THE EXPERIENCE COLUMN}
\label{tab:experience_performance}
\begin{tabular}{|c|c|c|c|c|}
\hline
Class & Precision & Recall & F1-Score & Support \\
\hline
Poor   & 0.95   & 1   & 0.97   & 727 \\
\hline
Fair   & 1   & 0.36   & 0.53   & 272 \\
\hline
Excellent  & 0.99  & 0.99  & 0.94  & 1214 \\
\hline
Accuracy &  &  &  & 0.91 \\
\hline
Hamming Loss &  &  &  & 0.0854 \\
\hline
\end{tabular}
\end{table}

Regarding the training dynamics, both the training and validation accuracy curves demonstrate continuous growth over 20 epochs, stabilizing at 91\% by the final epoch. The loss curves follow a similar trend, reducing from 1.0 to 0.2 for training loss and from 0.25 to 0.20 for validation loss, demonstrating that the model is generalizing well without overfitting.

\begin{figure}[htbp]
    \centering
    \includegraphics[width=0.5\textwidth]{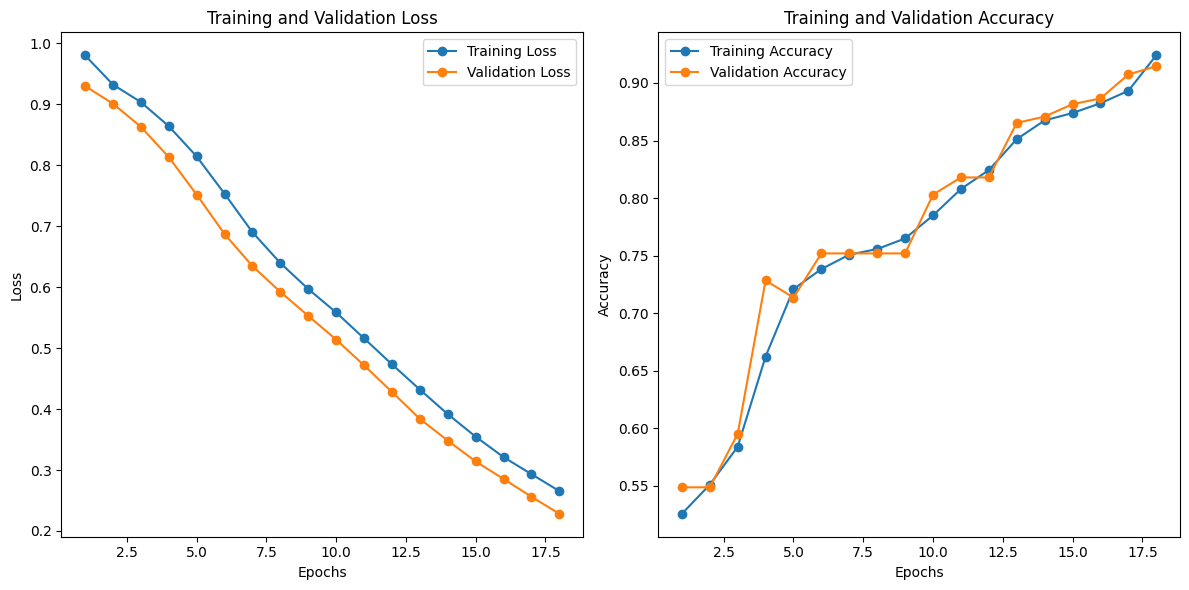}
    \caption{Training and validation loss with accuracy over epochs graph for Experience}
    \label{fig:wf1}
\end{figure}

The Fig.~\ref{fig:wf1} shows the training and validation loss with accuracy over epochs graph for the experience attribute.

\begin{figure}[htbp]
    \centering
    \includegraphics[width=0.35\textwidth]{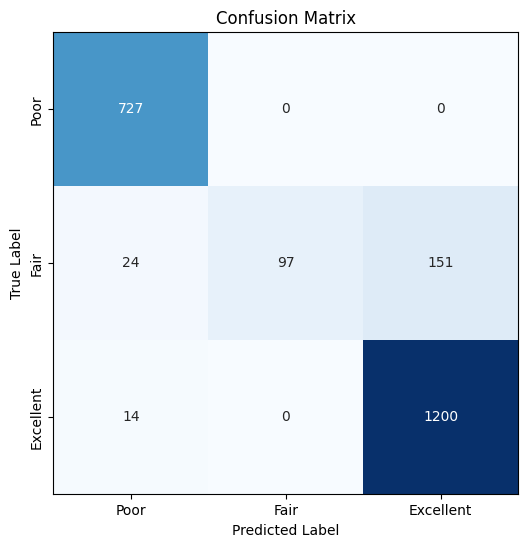}
    \caption{Model Prediction Results (Confusion Matrix for Experience)}
    \label{fig:wf1C}
\end{figure}

Analyzing the confusion matrix shown in the Fig~\ref{fig:wf1C}, all "Poor" samples were correctly classified; however, there were misclassifications in the "Fair" category, 97 samples were wrongly labeled as "Excellent," and 24 as "Poor." The "Excellent" category had low misclassifications, with only 14 samples misclassified as "Fair." Additionally, the model's Hamming Loss of 0.0854 illustrates its capacity to minimize errors effectively.

While the DistilRoBERTa model excels in most categories, boosting recall for the "Fair" category remains a possible area for additional optimization. Nonetheless, the model's high precision, recall, accuracy, and balanced loss patterns position it as a viable tool for proficiency score analysis tasks.

\subsection{Performance of DistilRoBERTa on Skills Attribute}

The DistilRoBERTa model attained an overall accuracy of 87\% in classifying the "Skills" column into the "Poor," "Fair," and "Excellent" categories, shown in the Table~\ref{tab:skill_performance}. For the "Poor" category, it performed dependably with a precision of 0.91, a recall of 0.86, and an F1-score of 0.88. In the "Fair" category, the model demonstrated excellent recall (0.89) but lower precision (0.77), leading to an F1-score of 0.83, which shows some overlap between classes. The "Excellent" category did the best with a precision of 0.98 and an F1-score of 0.92.

\begin{table}[h]
\centering
\caption{PERFORMANCE METRICS FOR THE SKILLS COLUMN}
\label{tab:skill_performance}
\begin{tabular}{|c|c|c|c|c|}
\hline
Class & Precision & Recall & F1-Score & Support \\
\hline
Poor   & 0.91   & 0.86   & 0.88   & 735 \\
\hline
Fair   & 0.77   & 0.89   & 0.83   & 752 \\
\hline
Excellent  & 0.98  & 0.87  & 0.92  & 726 \\
\hline
Accuracy &  &  &  & 0.87 \\
\hline
Hamming Loss &  &  &  & 0.1261 \\
\hline
\end{tabular}

\end{table}

\begin{figure}[htbp]
    \centering
    \includegraphics[width=0.5\textwidth]{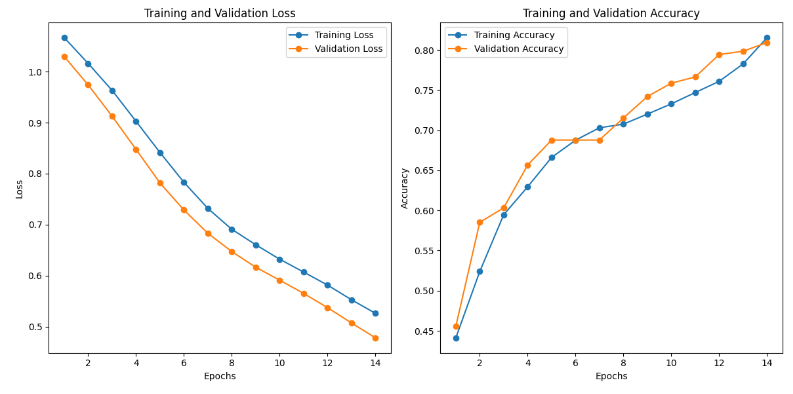}
    \caption{Training and validation loss with accuracy over epochs graph for Skills}
    \label{fig:wf2}
\end{figure}

The Fig~\ref{fig:wf2} shows the training and validation loss with accuracy over epochs graph for the skill attribute. Throughout 15 epochs, both training and validation accuracy steadily improved, ultimately stabilizing at 87\%, with no symptoms of overfitting. The loss curves displayed a consistent reduction, with training loss reducing from 1.0 to 0.4 and validation loss reaching 0.5, proving the model's capacity to generalize well while learning efficiently from the data.

\begin{figure}[htbp]
    \centering
    \includegraphics[width=0.35\textwidth]{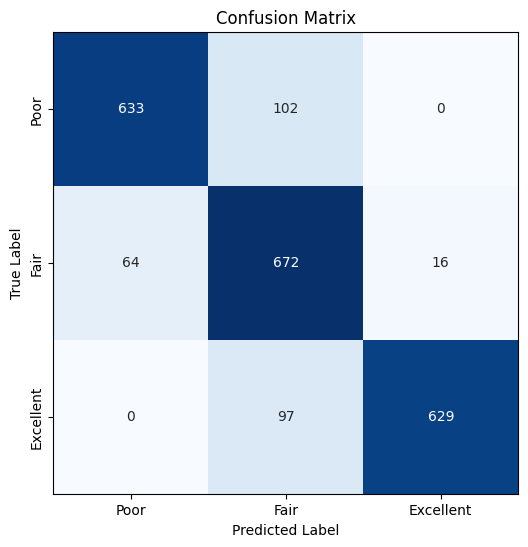}
    \caption{Model Prediction Results (Confusion Matrix for Skills)}
    \label{fig:wf2C}
\end{figure}

The confusion matrix shown in the Fig~\ref{fig:wf2C} revealed appropriate classification for most categories, but "Fair" samples were commonly misclassified as "Poor" or "Excellent." This overlap is further indicated by a Hamming Loss of 0.1261, suggesting space for improvement in discriminating between these closely related categories.

Overall, the model's performance is encouraging for multi-class sentiment categorization. While it displays outstanding optimization and resilient training dynamics, fine-tuning to better differentiate "Fair" from other classes could boost its precision and dependability for real-world applications.

\subsection{Performance of DistilRoBERTa on Education Attribute}

The DistilRoBERTa model fared well in classifying the "Education" column into "Poor," "Fair," and "Excellent" categories as displayed in the Table~\ref{tab:column_performance}. The "Poor" class achieved flawless results (Precision = 1.00, Recall = 1.00, F1-Score = 1.00), but both the "Fair" and "Excellent" groups scored an F1 of 0.87. Overall accuracy was 91\%, with a low Hamming Loss of 0.0886, indicating negligible misclassifications.

\begin{table}[h]
\centering
\caption{PERFORMANCE METRICS FOR THE EDUCATION COLUMN}
\label{tab:column_performance}
\begin{tabular}{|c|c|c|c|c|}
\hline
Class & Precision & Recall & F1-Score & Support \\
\hline
Poor   & 1   & 1   & 1   & 727 \\
\hline
Fair   & 0.89   & 0.87   & 0.83   & 751 \\
\hline
Excellent  & 0.85  & 0.87  & 0.92  & 735 \\
\hline
Accuracy &  &  &  & 0.91 \\
\hline
Hamming Loss &  &  &  & 0.0886 \\
\hline
\end{tabular}
\end{table}

\begin{figure}[htbp]
    \centering
    \includegraphics[width=0.5\textwidth]{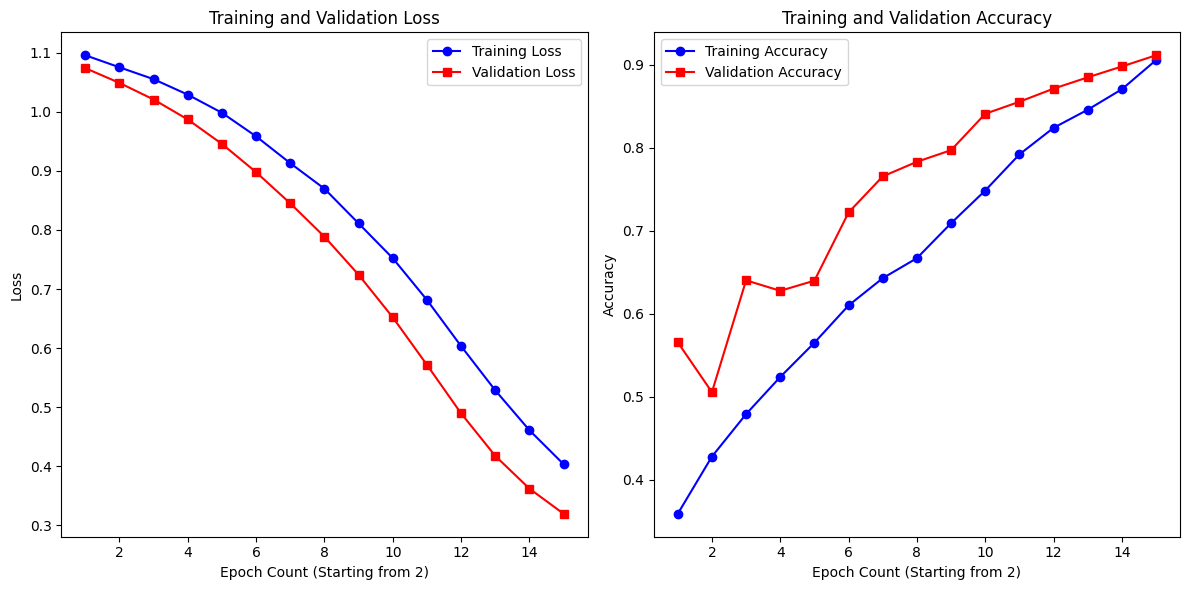}
    \caption{Training and validation loss with accuracy over epochs graph for Education}
    \label{fig:wf3}
\end{figure}

Training and validation accuracy grew steadily as shown in the Fig~\ref{fig:wf3}, reaching near-perfect levels, with validation accuracy stabilizing at 91\%. Training loss continually lowers, suggesting the model's improved capacity to minimize errors throughout training. Validation loss also reduces, illustrating the model's generalization capability across unknown data. By the final epochs, the alignment of training and validation loss curves further underscores the absence of overfitting, as both metrics approach a convergence point.

\begin{figure}[htbp]
    \centering
    \includegraphics[width=0.35\textwidth]{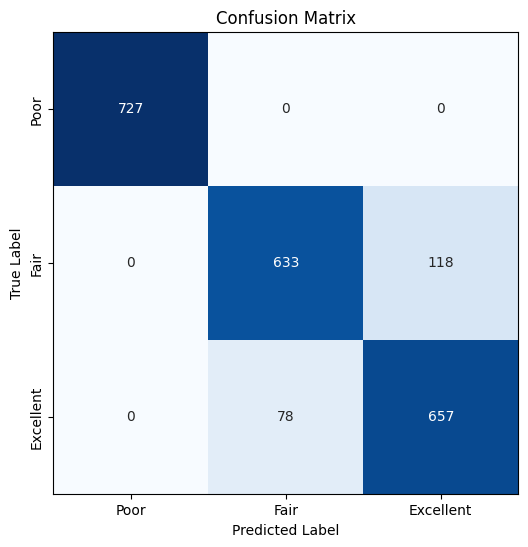}
    \caption{Model Prediction Results (Confusion Matrix for Education)}
    \label{fig:wf3C}
\end{figure}

The confusion matrix shown in the Fig~\ref{fig:wf3C} demonstrated faultless categorization of the "Poor" class, while some misclassifications occurred between "Fair" and "Excellent" (118 and 78 instances, respectively). These results demonstrate the model's strength in simpler categories and difficulty in discriminating between more similar classes

\subsection{Performance of DistilRoBERTa on About Attribute}

The DistilRoBERTa model achieves outstanding performance with a precision of 0.98, a recall of 0.99, and an F1 score of 0.99, exhibiting strong classification accuracy in the "Poor," "Fair," and "Excellent" categories as shown in the Table~\ref{tab:about_performance}. It has an overall accuracy of 0.99, indicating its trustworthiness for sentiment classification. 

\begin{table}[h]
\centering
\caption{PERFORMANCE METRICS FOR THE ABOUT COLUMN}
\label{tab:about_performance}
\begin{tabular}{|c|c|c|c|c|}
\hline
Class & Precision & Recall & F1-Score & Support \\
\hline
Poor   & 0.98   & 0.99   & 0.99   & 727 \\
\hline
Fair   & 0.99   & 0.98   & 0.99   & 751 \\
\hline
Excellent  & 0.99  & 0.99  & 0.99  & 735 \\
\hline
Accuracy &  &  &  & 0.99 \\
\hline
Hamming Loss &  &  &  & 0.0054 \\
\hline
\end{tabular}

\end{table}

\begin{figure}[htbp]
    \centering
    \includegraphics[width=0.35\textwidth]{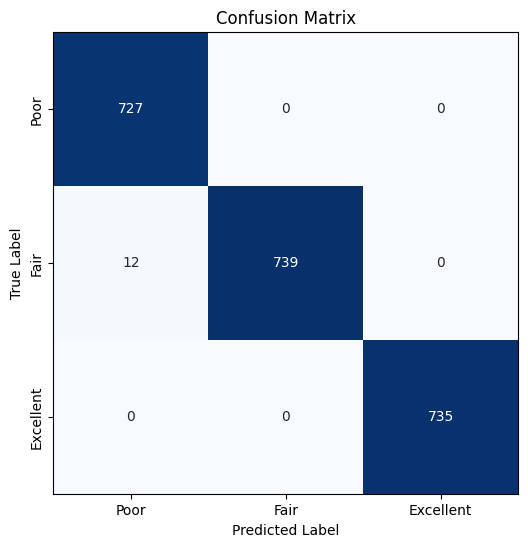}
    \caption{Model Prediction Results (Confusion Matrix of About Column)}
    \label{fig:wf4C}
\end{figure}

The confusion matrix shown in the Fig~\ref{fig:wf4C} indicates correct predictions for 727 "Poor," 739 "Fair," and 735 "Excellent" samples, with limited misclassifications, predominantly "Fair" as "Poor." 

\begin{figure}[htbp]
    \centering
    \includegraphics[width=0.5\textwidth]{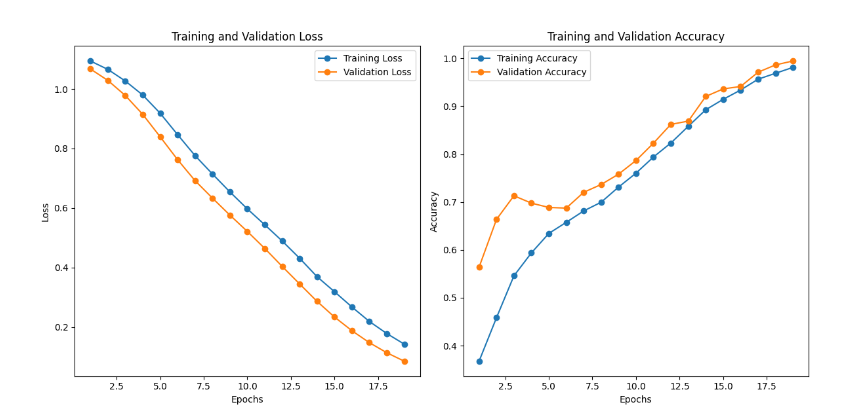}
    \caption{Training and validation loss with accuracy over epochs graph for About}
    \label{fig:wf4}
\end{figure}

Training and validation accuracy improve steadily, reaching approximately 91\% by the final epoch shown in the Fig~\ref{fig:wf4}. The loss curves reveal constant improvement, with training loss reducing from 1.0 to 0.2, and validation loss following a similar path. The Hamming loss of 0.0054 further underscores its low misclassifications rate.

Overall, the model’s high precision, recall, accuracy, and balanced loss trends show its robustness and usefulness for multi-class proficiency score classification tasks.


\subsection{Performance of DistilRoBERTa on Overall Attribute}

The Table~\ref{tab:overall_performance} presents a detailed assessment of the DistilRoBERTa-based model's performance on a multi-class educational categorization problem. The table displays the metrics for each class, displaying the model's ability to classify the Poor category with perfect precision, recall, and F1-score. For the Fair and Excellent categories, the F1 scores suggest balanced and consistent performance, with the model reaching an overall accuracy of 91\%.

\begin{table}[h]
\centering
\caption{PERFORMANCE METRICS FOR THE OVERALL COLUMN}
\label{tab:overall_performance}
\begin{tabular}{|c|c|c|c|c|}
\hline
Class & Precision & Recall & F1-Score & Support \\
\hline
Poor   & 1   & 1   & 1   & 735 \\
\hline
Fair   & 0.84   & 0.89   & 0.87   & 752 \\
\hline
Excellent  & 0.88  & 0.83  & 0.85  & 726 \\
\hline
Accuracy &  &  &  & 0.91 \\
\hline
\end{tabular}
\end{table}

\begin{figure}[htbp]
    \centering
    \includegraphics[width=0.5\textwidth]{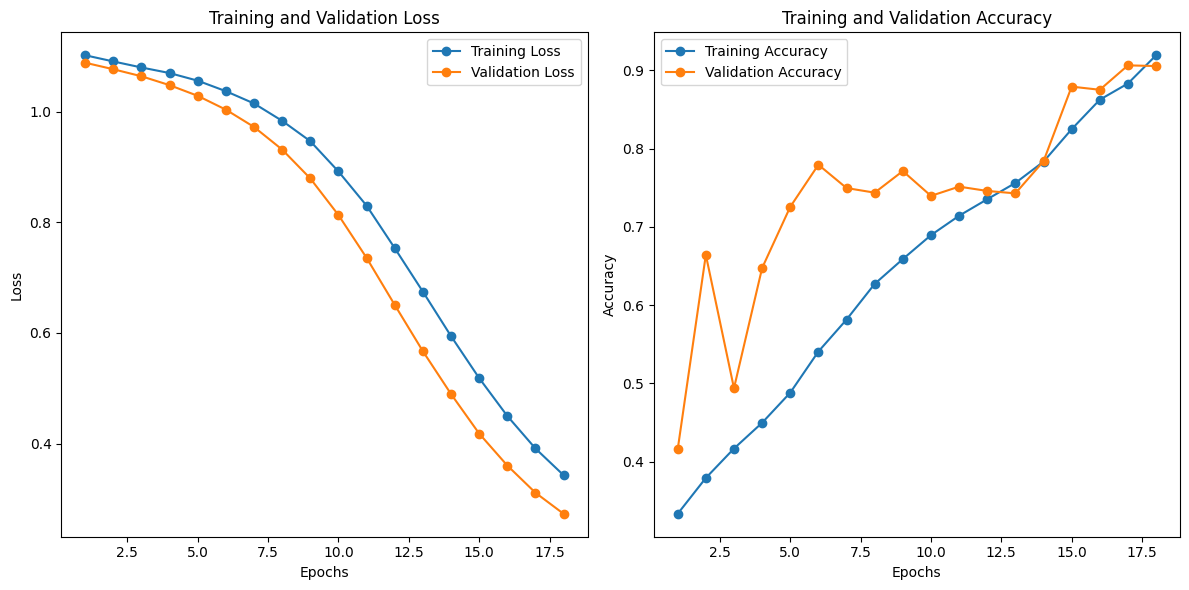}
    \caption{Training and validation loss with accuracy over epochs graph for Overall}
    \label{fig:wf5}
\end{figure}

The accompanying learning curves depict the growth of training and validation accuracy across 18 epochs, shown in the Fig~\ref{fig:wf5}. The consistent improvement in validation accuracy, which stabilizes at 91\%, indicates the model's robust learning capability and effective generalization without evidence of overfitting. The training loss starts at 1.1 and steadily drops to roughly 0.3, while the validation loss follows a similar pattern, converging at 0.35 by the final epochs. The convergence of training and validation loss by the last epochs demonstrates robust generalization and low overfitting. 

\begin{figure}[htbp]
    \centering
    \includegraphics[width=0.35\textwidth]{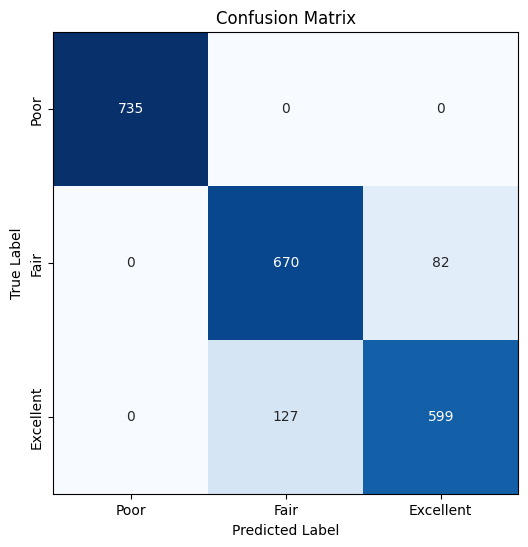}
    \caption{Model Prediction Results (Confusion Matrix for Overall)}
    \label{fig:wf5C}
\end{figure}

The confusion matrix, as shown in the Fig~\ref{fig:wf5C}, takes a closer look at the classification findings, highlighting the perfect categorization of the Poor category while identifying places for improvement in discriminating between Fair and Excellent. Misclassifications include 82 Fair cases predicted as Excellent and 127 Excellent cases predicted as Fair.

Together, these features provide a full evaluation of the model's capabilities, notably in basic classifications, and its problems in handling more complex differences, making it a good tool for examining classification performance in educational datasets.

\subsection{Ablation Study}
\label{sec:ablation}

An ablation study was conducted to evaluate the contribution of data augmentation, model fine-tuning, and the Fuzzy-TOPSIS decision layer. As shown in Table~\ref{tab:ablation}, the baseline DistilRoBERTa model trained on non-augmented data exhibits poor performance, indicating that the raw dataset alone is insufficient for reliable learning. The introduction of data augmentation and task-specific fine-tuning leads to a substantial improvement, increasing both overall F1-score and accuracy to 0.91 while significantly reducing the Hamming loss.

Using identical augmented LLM outputs, the integration of the Fuzzy-TOPSIS layer further improves ranking performance, increasing MAP from 0.93 to 0.99 and cosine similarity from 0.95 to 0.98. These results indicate that Fuzzy-TOPSIS contributes at the decision aggregation stage by explicitly modeling criterion importance and uncertainty, while leaving classification performance unchanged.

\begin{table}[htbp]
\centering
\caption{Ablation Study on the Impact of Augmentation, Fine-Tuning, and Fuzzy-TOPSIS}
\renewcommand{\arraystretch}{1.15}
\setlength{\tabcolsep}{4pt}
\begin{tabular}{|p{3.1cm}|c|c|c|c|c|}
\hline
\textbf{Method Variant} &
\textbf{F1} &
\textbf{Acc.} &
\textbf{HL} &
\textbf{MAP} &
\textbf{Cos. Sim.} \\ \hline

Baseline DistilRoBERTa \newline
(no aug., no TOPSIS)
& 0.16 & 0.27 & 0.73 & -- & -- \\ \hline

Augmented \& Fine-Tuned \newline
DistilRoBERTa (LLM-only)
& 0.91 & 0.91 & 0.094 & 0.93 & 0.95 \\ \hline

Augmented DistilRoBERTa + Fuzzy-TOPSIS
& 0.91 & 0.91 & 0.094 & \textbf{0.99} & \textbf{0.98} \\ \hline

\end{tabular}
\label{tab:ablation}
\end{table}

Table~\ref{tab:ablation_llm} evaluates the impact of data augmentation and knowledge distillation on classification performance. The baseline DistilRoBERTa model trained on non-augmented data performs poorly, indicating that the limited and imbalanced raw dataset does not provide sufficient semantic coverage for effective learning. After applying data augmentation and fine-tuning, performance improves substantially, with the overall F1-score increasing from 0.16 to above 0.90 and a corresponding reduction in Hamming loss.

Among the augmented models, distilled architectures (DistilRoBERTa and LastBERT) outperform the full RoBERTa-base model, particularly in terms of F1-score and error rate, indicating superior generalization under limited-data conditions. Although LastBERT achieves marginally higher overall performance, DistilRoBERTa was selected as the backbone model due to its favorable balance between accuracy, architectural simplicity, and compatibility with downstream fuzzy decision aggregation. Its more stable aspect-wise predictions and lower variance across criteria make it particularly suitable for integration with multi-criteria decision-making methods such as Fuzzy-TOPSIS.

\begin{table}[htbp]
\centering
\caption{Ablation Study on Data Augmentation and Model Architecture}

\renewcommand{\arraystretch}{1.15}
\setlength{\tabcolsep}{4pt}
\begin{tabular}{|p{2.6cm}|c|c|c|c|}
\hline
\textbf{Model Variant} &
\textbf{Params (M)} &
\textbf{F1} &
\textbf{Accuracy} &
\textbf{Hamming Loss} \\ \hline

Baseline DistilRoBERTa \newline
(no augmentation)
& 82 & 0.16 & 0.27 & 0.73 \\ \hline

Augmented \& Fine-Tuned \newline
DistilRoBERTa
& 82 & 0.91 & 0.91 & 0.094 \\ \hline

Augmented \& Fine-Tuned \newline
LastBERT (distilled)
& 29 & 0.93 & 0.93 & 0.0718 \\ \hline

Augmented \& Fine-Tuned \newline
RoBERTa-base (full)
& 125 & 0.85 & 0.85 & 0.1509 \\ \hline

\end{tabular}
\label{tab:ablation_llm}
\end{table}


The performance evaluation of the TOPSIS-based ranking system was conducted using both human-generated and LLM-generated proficiency scores. The following analyses highlight the consistency and effectiveness of the proposed ranking framework.
\subsection{Ranking Performance Using Human-Generated Proficiency Scores}

The results from the human-generated proficiency scores indicate the usefulness of the TOPSIS methodology. The Root Mean Square Error (RMSE) of 0.043 indicates a minimal margin of error, verifying the dependability of the ranking method. Similarly, the Mean Absolute Error (MAE) of 0.036 underlines the smallest divergence from the optimal ranks. The Cosine Similarity score of 0.981 demonstrates a strong alignment with the optimal solution, verifying the accuracy of the rankings.
Furthermore, the Normalized Discounted Cumulative Gain (NDCG) score of 0.911 verifies the ranking quality and relevance of the candidates. The Mean Average Precision (MAP) and Mean Reciprocal Rank (MRR) both earned scores of 0.999, indicating the model's ability to rank relevant candidates well. These indicators together verify that the TOPSIS-based system operates with outstanding consistency and precision in evaluating human-generated competency results.

\subsection{Ranking Performance Using DistilRoBERTa-Generated Scores with Fuzzy-TOPSIS}

The performance metrics acquired from the LLM model-generated competency scores further prove the robustness of the TOPSIS technique. The Root Mean Square inaccuracy (RMSE) and Mean Absolute Error (MAE) remain stable at 0.043 and 0.036, respectively, demonstrating similar levels of inaccuracy as reported with human-generated scores. The Cosine Similarity score of 0.983 significantly surpasses that of the human-generated scores, suggesting a marginally improved alignment with the ideal solution.
Additionally, the Normalized Discounted Cumulative Gain (NDCG) score climbed to 0.926, reflecting an advance in ranking quality and relevance. The Mean Average Precision (MAP) and Mean Reciprocal Rank (MRR), both exceeding 0.999, continue to highlight the TOPSIS system's potential to select relevant candidates with outstanding efficiency. These findings underline the possibility of LLM-generated proficiency scores in keeping or even surpassing the performance observed with human-generated scores. This indicates the possibility of employing LLMs for automated staff selection activities, giving a scalable and consistent alternative to human judgments. It should be noted that cosine similarity is employed as a descriptive agreement measure to assess alignment between human and LLM–Fuzzy TOPSIS rankings. Given the deterministic nature of the ranking outputs and the limited number of candidates, no claim of statistical significance is made regarding the marginal difference between the reported cosine similarity values.

\subsection{Comparison of Human-TOPSIS and LLM--Fuzzy TOPSIS Rankings}

The model's TOPSIS scores closely paralleled human assessments, suggesting its ability to effectively represent the relative performance of candidates. While there were slight deviations, the main trend was constant. The model's ranks were significantly associated with human rankings. Both algorithms consistently placed Candidate Raju Ahmed at the top position, indicating the model's capacity to identify top-performing applicants. These results imply that the model can be a beneficial tool in candidate evaluation, giving consistent and trustworthy assessments.

\begin{figure}[htbp]
    \centering
    \includegraphics[width=0.5\textwidth]{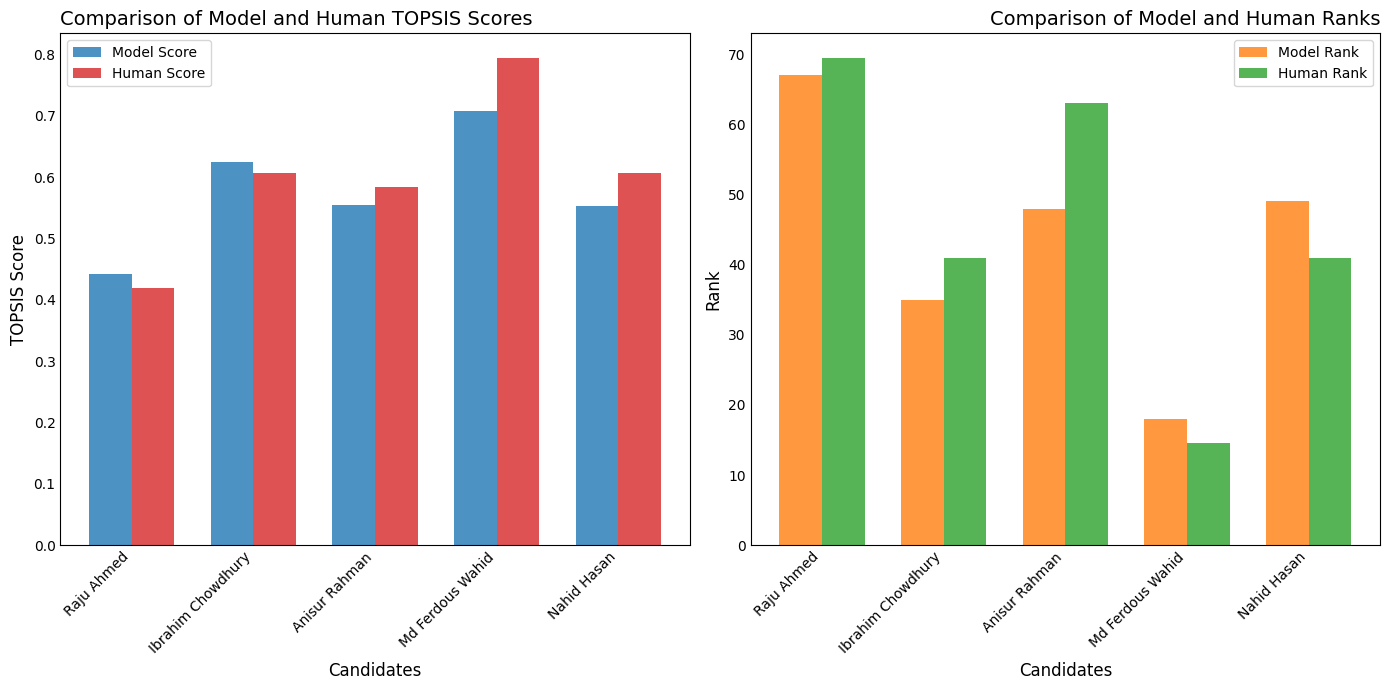}
    \caption{Comparison of candidate scores and rankings produced by Human-TOPSIS and LLM–Fuzzy TOPSIS. The left panel shows TOPSIS closeness scores, while the right panel compares the final candidate ranks.}
    \label{fig:wf}
\end{figure}

\begin{table}[htbp]
\centering
\caption{Performance Metrics for the TOPSIS Method}
\begin{tabular}{|c|c|}
\hline
\textbf{Metric} & \textbf{Value} \\ \hline
MAP & 0.99 \\ \hline
NDCG & 0.9113 \\ \hline
MRR & 0.99 \\ \hline
RMSE & 0.04392 \\ \hline
MAE & 0.0368 \\ \hline
Cosine Similarity & 0.9819 \\ \hline
\end{tabular}
\label{tab:topsis_results}
\end{table}

\begin{table}[h!]
\centering
\caption{Rankings Evaluation}
\resizebox{\columnwidth}{!}{%
\begin{tabular}{|l|c|c|c|}
\hline
\textbf{Candidates} & \shortstack{\textbf{Ranking by} \\ \textbf{Human-TOPSIS}} & \shortstack{\textbf{Ranking by} \\ \textbf{LLM--Fuzzy TOPSIS}} & \shortstack{\textbf{Ranking by} \\ \textbf{Human Experts}} \\
\hline
A. Abu Sifat & 1 & 3 & 2 \\
\hline
B. Sheikh Shofiullah & 2 & 1 & 3 \\
\hline
C. Najmul Hoq & 3 & 2 & 4 \\
\hline
D. Anjum Haz & 4 & 4 & 5 \\
\hline
E. Rajesh Debnath & 7 & 5 & 1 \\
\hline
F. Mohammad Asif & 6 & 6 & 6 \\
\hline
G. Md. Naushad & 8 & 8 & 7 \\
\hline
H. Towhid Zaman & 9 & 9 & 8 \\
\hline
I. Miraz Anik & 10 & 13 & 9 \\
\hline
J. Kazi Nasir & 15 & 15 & 10 \\
\hline
\end{tabular}%
}
\label{table:rankings}
\end{table}


\section{Discussion}
\label{sec:discussion}

This study demonstrates the efficacy of transformer-based models and the proposed LLM-TOPSIS framework in classifying and ranking candidate profiles. The multi-class classification model achieved a strong overall accuracy of 91\%, with F1-scores of 1.00, 0.87, and 0.85 for the \textit{Poor}, \textit{Fair}, and \textit{Excellent} classes, respectively. These scores, along with the confusion matrix, confirm the model’s effectiveness in distinguishing candidate quality levels with minimal misclassifications. Such performance validates the potential of fine-tuned transformer architectures for nuanced eligibility classification based on textual data.

Beyond classification, the LLM-TOPSIS method was evaluated against both Human-TOPSIS and direct Human rankings. The generated rankings showed a high level of alignment with human evaluators across multiple positions. For instance, \textit{Abu Sifat} was consistently ranked among the top candidates by all three methods. The rankings for \textit{Sheikh Shofiullah} and \textit{Najmul Hoq} also closely matched across LLM-TOPSIS and Human scores, indicating that the LLM’s evaluations are well-calibrated to human reasoning patterns.

Quantitative metrics support this alignment: Mean Average Precision (MAP) and Mean Reciprocal Rank (MRR) both scored 0.99, cosine similarity reached 0.98, and the low RMSE (0.0439) and MAE (0.0368) further confirm that the LLM-TOPSIS framework can closely approximate expert judgment with minimal error. These results suggest that the model not only replicates human rank order effectively but also does so with statistical consistency and reliability.

The ablation analysis further confirms that while data augmentation and fine-tuning significantly enhance classification performance, the integration of Fuzzy-TOPSIS contributes additional gains at the decision-making level, improving ranking quality without altering classification accuracy.

Overall, the integration of transformer-based classification and the LLM-TOPSIS ranking mechanism provides a robust, interpretable, and scalable system for candidate evaluation. It achieves both high classification accuracy and rank order fidelity, making it suitable for real-world applications where fairness, transparency, and human-aligned decision-making are essential, such as academic admissions, hiring, and scholarship selection.

\section{Limitations and Future Work}
\label{sec:limitations_future}

Although the proposed LLM-TOPSIS framework achieved strong classification and ranking performance, several limitations remain that open avenues for future research. The present study utilized approximately 100 LinkedIn profiles focused on software engineering roles. While this curated dataset effectively demonstrated methodological feasibility, its limited scope may restrict generalizability across industries and geographic regions. The relatively small dataset size primarily stems from the manual and labor-intensive process of hand-scoring and labeling each profile by domain experts across multiple evaluation criteria. Although this constrained the number of samples, it ensured label quality, reliability, and consistent annotation—an essential aspect when validating a multi-criteria decision-making framework that depends on subjective human evaluation.

Future work will focus on expanding the dataset to multiple professional domains, including healthcare, finance, and education, to enable cross-domain validation and to mitigate potential selection bias. Incorporating a larger and more diverse dataset will further strengthen the model’s robustness and fairness assessment across varying demographic and regional distributions. Additionally, we plan to integrate automated or semi-automated annotation pipelines using large language models to scale the labeling process while maintaining expert-level consistency.

Another key direction involves refining the fuzzy weight learning process using data-driven entropy optimization or adaptive fuzzy-AHP formulations to dynamically adjust criterion importance across contexts. Moreover, interpretability and explainability remain critical for real-world deployment. Future studies will employ frameworks such as SHAP or LIME to visualize LLM attention and illustrate the contribution of textual segments to each decision score. Cross-cultural and linguistic analyses will also be conducted to examine how regional or language-specific variations in professional profiles affect ranking outcomes.

In addition, future work will include controlled robustness experiments such as noise injection, profile truncation, and perturbation of textual attributes to quantitatively assess sensitivity to profile quality variations.

Overall, these extensions aim to improve scalability, generalization, and transparency, ensuring that the proposed LLM-TOPSIS framework continues to align with ethical and practical requirements in AI-driven personnel evaluation.


\section{Conclusion}
\label{sec:conclusion}

In conclusion, this research emphasizes the promise of transformer-based models in automating people selection, particularly for analyzing software engineering characteristics. By integrating modern natural language processing techniques, the system illustrates the viability of scalable and consistent analysis, minimizing reliance on old human methodologies. The results show that such models can streamline recruiting processes, maintaining efficiency while eliminating biases inherent in human evaluations. Future initiatives include refining the models with different datasets, including complementary data sources like resumes, and utilizing explainability tools to build trust and transparency. This study serves as a stepping stone toward incorporating AI-driven solutions in human resource management, opening the door for better-informed and objective decision-making in recruitment procedures. Additionally, implementing the models in real-world contexts and integrating explainability approaches like SHAP or LIME would boost model transparency and test their usefulness in practice. In summary, this research establishes the framework for automated personnel selection utilizing state-of-the-art NLP techniques, contributing to more data-driven and unbiased decision-making in human resource management.

Overall, this work demonstrates that combining LLM-based textual understanding with fuzzy multi-criteria decision-making yields a transparent, scalable, and human-aligned framework for automated personnel selection.

\appendices
\section{Expert Evaluation Rubric}
\label{appendix:rubric}

This appendix presents the scoring rubric collaboratively developed by three domain experts and employed to label each LinkedIn profile. Each criterion—\textit{Experience}, \textit{Skills}, \textit{Education}, and \textit{About}. This was evaluated on a five-point Likert scale (1–5), where higher scores indicate stronger professional relevance. A single expert annotated all profiles to ensure intra-rater consistency, while the rubric itself was validated by the other two experts before annotation.

\begin{table*}[htbp]
\centering
\caption{Expert Evaluation Rubric for Candidate Assessment}
\renewcommand{\arraystretch}{1.15}
\begin{tabular}{p{1cm} p{2.0cm} p{2.8cm} p{2.8cm} p{2.8cm} p{3.0cm}}
\hline
\textbf{Score} & \textbf{Meaning} & \textbf{Experience} & \textbf{Skills} & \textbf{Education} & \textbf{About / Summary} \\
\hline
1 (Poor) & Significantly below expectations & No relevant professional or internship experience & No demonstrable or unrelated skills & Incomplete or irrelevant education & Lacks clarity or professional focus \\

2 (Fair) & Below average & Minimal relevant experience (<1 yr) & Basic exposure to tools; lacks depth & Some relevant coursework but weak record & Generic or vague description \\

3 (Moderate) & Meets minimum expectations & 1–2 yrs relevant experience or internships & Adequate foundational skills with limited projects & Completed Bachelor's in computing or related field & Basic articulation of achievements and goals \\

4 (Good) & Above average & 2–5 yrs progressive, relevant experience & Strong skills across multiple frameworks and tools & Bachelor's or Master's degree in relevant field & Well-structured, coherent professional summary \\

5 (Excellent) & Outstanding / industry-ready & 5+ yrs experience, leadership or specialization & Advanced technical expertise and certifications & Advanced or distinguished academic qualification & Persuasive, confident summary demonstrating excellence \\
\hline
\end{tabular}
\end{table*}

\noindent
The rubric provided a standardized evaluation framework linking numeric scores to qualitative judgments, forming the basis for consistent and reproducible labeling in the proposed model.


\bibliography{references} 

\begin{IEEEbiography}[{\includegraphics[width=1in,height=1.25in,clip,keepaspectratio]{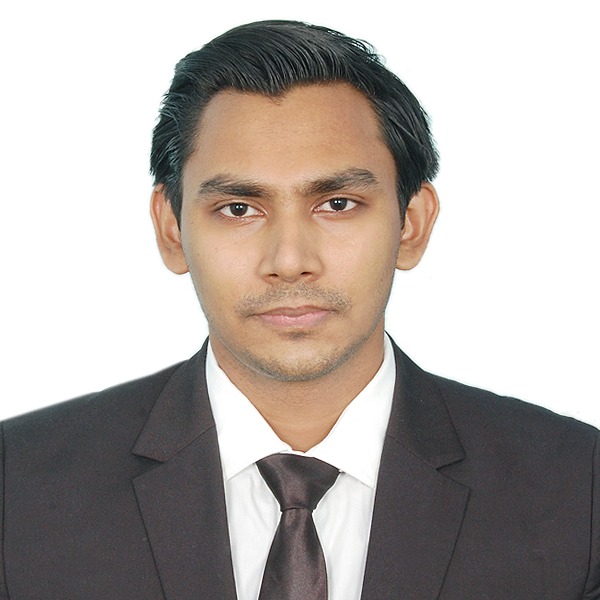}}]{Shahria Hoque} received his M.Sc. in Computer Science and Engineering from BRAC University, Bangladesh. His research interests lie primarily in the areas of Artificial Intelligence, Natural Language Processing (NLP), and Human-Computer Interaction (HCI). He is particularly focused on the application of deep learning techniques in sentiment analysis, text classification, and user-centered system design. He has experience in developing intelligent systems that enhance user engagement and improve accessibility in digital interfaces.

His research has been presented at the IEEE International Conference on Computer and Information Technology (ICCIT), where he authored a peer-reviewed paper. He continues to explore interdisciplinary intersections of AI with cognitive science and user behavior modeling.
\end{IEEEbiography}

\begin{IEEEbiography}[{\includegraphics[width=1in,height=1.25in,clip,keepaspectratio]{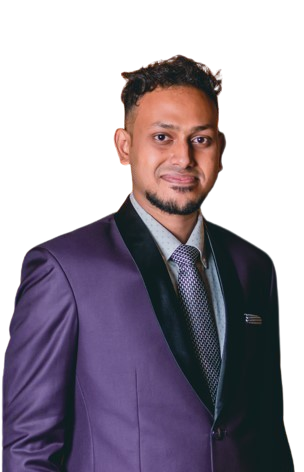}}]{Ahmed Akib Jawad Karim} received his B.Sc. degree in Computer Science and Engineering from North South University, Dhaka, Bangladesh, where he graduated with \textit{summa cum laude}. He is currently working as a lecturer in the Department of Computer Science and Engineering at BRAC University, Dhaka. His major field of study includes natural language processing, computer vision, and machine learning. He has prior experience as an undergraduate and graduate teaching assistant at North South University. His teaching responsibilities include courses in Programming languages (Java, Python), Artificial Intelligence, Data Structures, Algorithms, Database Systems, Software Engineering, and Systems Analysis. He is actively engaged in creating course content, supervising labs, and mentoring students in research.

Mr. Karim has authored multiple research articles in the Natural Language Processing and Computer Vision domains. His current research focuses on lightweight transformer architectures, open-set learning, and clinical text classification. He has authored and co-authored several research papers, which are published in reputed Q1 journals such as \textit{PLOS Computational Biology} and \textit{PLOS ONE}, as well as IEEE-affiliated conferences like ICCIT. Mr. Karim's long-term goals include contributing to interdisciplinary research and improving the quality of undergraduate CS education through innovative teaching and mentoring.

\end{IEEEbiography}

\begin{IEEEbiography}[{\includegraphics[width=1in,height=1.25in,clip,keepaspectratio]{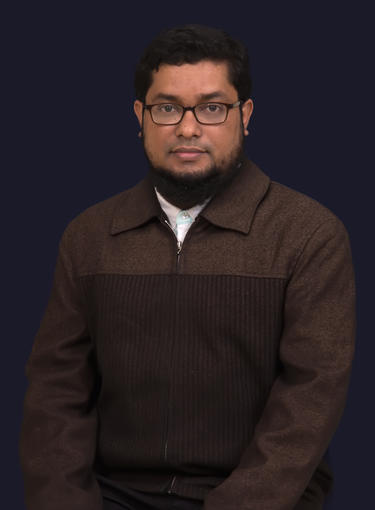}}]{Md. Golam Rabiul Alam} is a Professor in the Department of Computer Science and Engineering at BRAC University, Dhaka, Bangladesh. He received his Ph.D. in Computer Engineering from Kyung Hee University, South Korea, in 2017. He also served as a Postdoctoral Researcher in the same department from March 2017 to February 2018. He holds a B.S. in Computer Science and Engineering from Khulna University and an M.S. in Information Technology from the University of Dhaka, Bangladesh.

Dr. Alam has published more than 150 research articles in reputed journals and conference proceedings. He is the holder of three registered patents in the areas of ambient assisted living, mobile cloud computing, and mobile fog computing. He actively serves as a reviewer for numerous prestigious venues, including \textit{IEEE Communications Magazine}, \textit{IEEE Transactions on Network and Service Management}, \textit{IEEE Access}, \textit{Elsevier Vehicular Communications}, \textit{Elsevier Journal of Systems Architecture}, and several international conferences such as IEEE/IFIP IM, NOMS, IEEE ICC, APNOMS, ICOIN, ECCE, ACM ICUIMC (IMCOM), and ICCA (UAE).

His research interests include Healthcare-IoT networks, mental-health informatics, ambient intelligence systems, mobile-cloud and edge computing, affective computing, and UAV image processing. Dr. Alam is a member of the IEEE Computer Society, Consumer Electronics Society, and Industrial Electronics Society. He has received several best paper awards at national and international research conferences.
\end{IEEEbiography}

\begin{IEEEbiography}[{\includegraphics[width=1in,height=1.25in,clip,keepaspectratio]{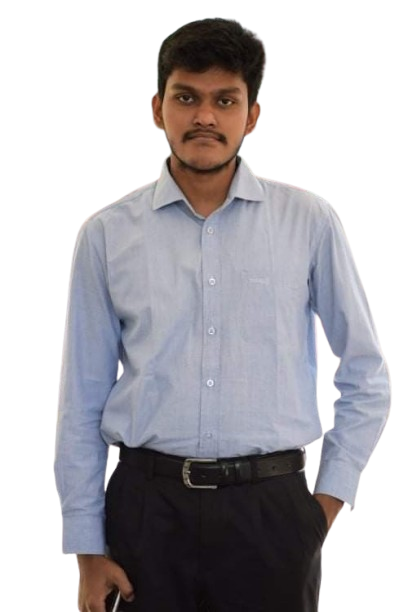}}]{Nirjhar Gope} received his B.Sc. in Computer Science and Engineering from BRAC University, Bangladesh. His research interests span machine learning, natural language processing (NLP), and the application of artificial intelligence in healthcare and linguistics. He is particularly focused on building data-driven systems that leverage large language models and deep learning for real-world problem-solving.

He has co-authored peer-reviewed work and contributed to several research projects involving intelligent systems, medical informatics, and language modeling. His goal is to develop AI solutions that bridge the gap between human language understanding and practical applications across diverse domains.
\end{IEEEbiography}

\EOD

\end{document}